\documentclass[twoside,11pt]{article}
\pdfoutput=1
\usepackage{jmlr2e}

\usepackage{amssymb}
\usepackage{lineno}

\usepackage[utf8]{inputenc} % allow utf-8 input
\usepackage[T1]{fontenc}    % use 8-bit T1 fonts
\usepackage{hyperref}
\usepackage{url}
\usepackage{pstricks,pst-node,pst-tree,pstricks-add}
\usepackage{booktabs}       % professional-quality tables
\usepackage{amsfonts}       % blackboard math symbols
\usepackage{nicefrac}       % compact symbols for 1/2, etc.
\usepackage{microtype}      % microtypography
\usepackage{xcolor}
\usepackage{tikz}
\usepackage{gensymb}

\usepackage{graphicx}
\usepackage{amsmath,amsfonts,bm}
\usepackage{rotating}
\usepackage[toc,page]{appendix}

\newcommand{\indep}{\rotatebox[origin=c]{90}{$\models$}}

\jmlrheading{1}{2018}{1-48}{4/00}{10/00}{Chaochao Lu, Bernhard Sch\"olkopf and Jos\'e Miguel Hern\'andez-Lobato}

\ShortHeadings{DRL in Observational Settings}{Lu, Sch\"olkopf and Hern\'andez-Lobato}
\firstpageno{1}

\begin{document}

\title{Deconfounding Reinforcement Learning \\in Observational Settings}

\author{\name Chaochao Lu \email cl641@cam.ac.uk \\
       \addr
       Department of Engineering\\
       University of Cambridge \\
       Cambridge, United Kingdom \vspace{2mm}\\ 
       Department of Empirical Inference \\
       Max Planck Institute for Intelligent Systems \\
       T\"ubingen, Germany
       \AND
       \name Bernhard Sch\"olkopf \email 
       bs@tuebingen.mpg.de\\
       \addr 
       Department of Empirical Inference \\
       Max Planck Institute for Intelligent Systems \\
       T\"ubingen, Germany
       \AND
       Jos\'e Miguel Hern\'andez-Lobato \email
       jmh233@cam.ac.uk \\
       \addr
       Department of Engineering\\
       University of Cambridge \\
       Cambridge, United Kingdom \vspace{2mm}\\ 
       Microsoft Research \\
       Cambridge, United Kingdom \vspace{2mm}\\ 
       Alan Turing Institute \\
       London, United Kingdom 
       }

\editor{}

\maketitle

\begin{abstract}
We propose a general formulation for addressing reinforcement learning (RL) problems in settings with observational data. That is, we consider the problem of learning good policies solely from historical data in which unobserved factors (confounders) affect both observed actions and rewards. Our formulation allows us to extend a representative RL algorithm, the Actor-Critic method, to its deconfounding variant, with the methodology for this extension being easily applied to other RL algorithms. 
In addition to this, we develop a new benchmark for evaluating deconfounding RL algorithms by modifying the OpenAI Gym environments and the MNIST dataset. Using this benchmark, we demonstrate that the proposed algorithms are superior to traditional RL methods in confounded environments with observational data. To the best of our knowledge, this is the first time that confounders are taken into consideration for addressing full RL problems with observational data. Code is available at https://github.com/CausalRL/DRL.
\end{abstract}

\section{Introduction}
In recent years, much attention has been devoted to the development of reinforcement learning (RL) algorithms with the goal of improving treatment policies in healthcare \citep{gottesman2018evaluating}. RL algorithms have been proposed to infer better decision-making strategies for mechanical ventilation \citep{prasad2017reinforcement}, sepsis management \citep{raghu2017deep,raghu2017continuous}, and treatment of schizophrenia \citep{shortreed2011informing}. In healthcare, a common practice is to focus on the observational setting in which we learn policies solely from historical data produced by real environments, instead of learning policies by actively taking actions in the traditional RL setting. The reason for this is that we do not wish to experiment with patients' lives without evidence that the proposed treatment strategy is better than the current practice \citep{gottesman2018evaluating}.\footnote{Another example is in finance where, due to costs in terms of time and money, it is often impractical to evaluate trading strategies by actually buying and selling stocks in the market.} As pointed out by \citet{raghu2017deep}, RL also has advantages over other machine learning algorithms even in the observational setting, especially in two situations: when the optimal treatment strategy is unclear in medical literature \citep{marik2015demise}, and when training examples do not represent optimal behavior. 

Another approach that has been explored and used is causal inference \citep{pearl2009causality}. Causal inference is a fundamental notion in science and plays an increasingly important role in healthcare and medicine \citep{liuusing,soleimani2017treatment,schulam2017reliable,alaa2017deep,alaa2018bayesian,atan2016constructing}. Especially in the observational setting, causal inference offers a powerful tool to deal with confounding that occurs when a hidden variable influences both the treatment and the outcome \citep{pearl2018bookofwhy}. It is widely acknowledged that confounding is the most crucial aspect of inferring the effect of the treatment on the outcome from observational data \citep{louizos2017causal}, because not taking confounding into account might lead to choosing a wrong treatment. However, most approaches in causal inference to tackling confounding are restricted to fixed treatments which do not vary over time \citep{louizos2017causal}. When treatments are allowed to vary over time, the space of combinational treatment strategies increases exponentially \citep{peters2017elements,hernan2018causal}. In this case, it is still unclear how to apply causal inference to dealing with confounding in such sequential data where treatments changes over time. 

On the basis of the discussion above, in this paper we attempt to combine advantages of RL and causal inference to cope with an important family of RL problems in the observational setting, that is, learning good policies solely from the historical data produced by real environments with \textit{confounding bias}. This type of problem will become increasingly common in future RL research with the burgeoning development in healthcare and medicine. To the best of our knowledge, however, little work has been done in this promising area of integrating RL with causal inference \citep{bareinboim2015bandits,forney2017counterfactual,buesing2018woulda}. In contrast, confounders have been extensively studied in epidemiology, sociology, and economics. Take for example the widespread kidney stones in which the size of the kidney stone is a confounding factor affecting both the treatment and the recovery \citep{peters2017elements,pearl2009causality}. Correcting for the confounding effect of the size of the kidney stone is crucial for determining how to choose an effective treatment. Similarly, in RL, if unobserved potential confounders exist, they affect both actions and rewards as an agent interacts with environments, and eventually influence the policy to be optimized. 

Let us for a moment stick with the example of the kidney stones and assume that physicians need to take a series of steps to treat this disease. We further assume that, during the course of treatment, physicians have natural predilections of the treatment choices as a function of the stone size. That is, it is more likely for them to choose some treatment when patients have large stones and to choose another treatment when patients have small stones. In the observational setting, given only such a set of historical records about physicians' treatments and outcomes on patients, it is extremely challenging or even impossible to learn an optimal treatment policy due to the existence of the confounder (i.e., the size of kidney stones). 

To this end, we present a general formulation for addressing RL problems with observational data, namely \textit{deconfounding reinforcement learning} (DRL). More specifically, given several common confounding assumptions, we first estimate a latent-variable model from observational data. Under some conditions for identification, through the latent-variable model we can simultaneously discover the latent confounders and infer how they affect actions and reward. Then the confounders in the model can be adjusted for, using the causal language developed by \cite{pearl2009causality}, and finally we optimize the policy based on the resulting deconfounding model. 

On the basis of the proposed formulation, we extend one popular RL algorithm, the Actor-Critic method, to its corresponding deconfounding variant. Note that our procedure for obtaining this deconfounding variant can be easily applied to other RL algorithms. Due to lack of datasets in this respect, we revise the classic control toolkit in OpenAI Gym \citep{openaigym}, making it a benchmark for comparison of DRL algorithms. In addition, we also devise a confounding version of the MNIST dataset \citep{lecun1998gradient} to verify the performance of our causal model. Finally, we conduct extensive experiments to demonstrate the superiority of the proposed formulation compared to traditional RL algorithms in confounded environments with observational data.  

To sum up, our contributions in this paper are as follows:
\begin{enumerate}
\item We propose a general formulation for addressing RL problems in confounded environments with observational data, namely \textit{deconfounding reinforcement learning} (DRL);
\item We present the deconfounding variant of Actor-Critic methods, obtained through a methodology that can be easily applied to other RL methods;
\item We develop a benchmark for DRL by revising the toolkit for classic control in OpenAI Gym \citep{openaigym} and by devising a version of the MNIST dataset with confounders \citep{lecun1998gradient};
\item We perform a comprehensive comparison of our DRL algorithm with its vanilla version, showing that the proposed approach has an advantage in confounded environments.
\item To the best of our knowledge, this is the first attempt to build a bridge between confounding and the full RL problem. This is one of few research papers aiming at understanding the connections between causal inference and the full RL setting.
\end{enumerate}

\section{Background}

In this section, we briefly review \textit{confounding} in causal inference. We recommend Pearl's excellent monograph for further reading \citep{pearl2009causality}.

\subsection{Simpson's Paradox}
Let us begin with an example of one of the most famous paradoxes in statistics: Simpson's Paradox. Consider the previously mentioned kidney stones  \citep{peters2017elements}. We collect electronic patient records to investigate the effectiveness of two treatments against kidney stones. Although the overall probability of recovery is higher for patients who took treatment $b$, treatment $a$ performs better than treatment $b$ both on patients with small kidney stones and with large kidney stones. More precisely, we have 
\begin{align}
p(R=1|T=b) &> p(R=1|T=a); \qquad \text{but} \nonumber \\
  p(R=1|T=b,Z=l) &< p(R=1|T=a,Z=l), \nonumber \\
p(R=1|T=b,Z=s) &< p(R=1|T=a,Z=s);
\end{align}
where $Z$ is the size (large or small) of the stone, $T$ the treatment, and $R$ the recovery (all binary). How do we cope with this change in conclusions? Which treatment do you prefer if you had kidney stones? Does treatment $b$ cause recovery? As described in the following section, the answers to these questions depend on the causal relationship between treatment, recovery, and the size of the kidney stone.

\begin{figure}[t]
\centering
\includegraphics[width=0.3\linewidth]{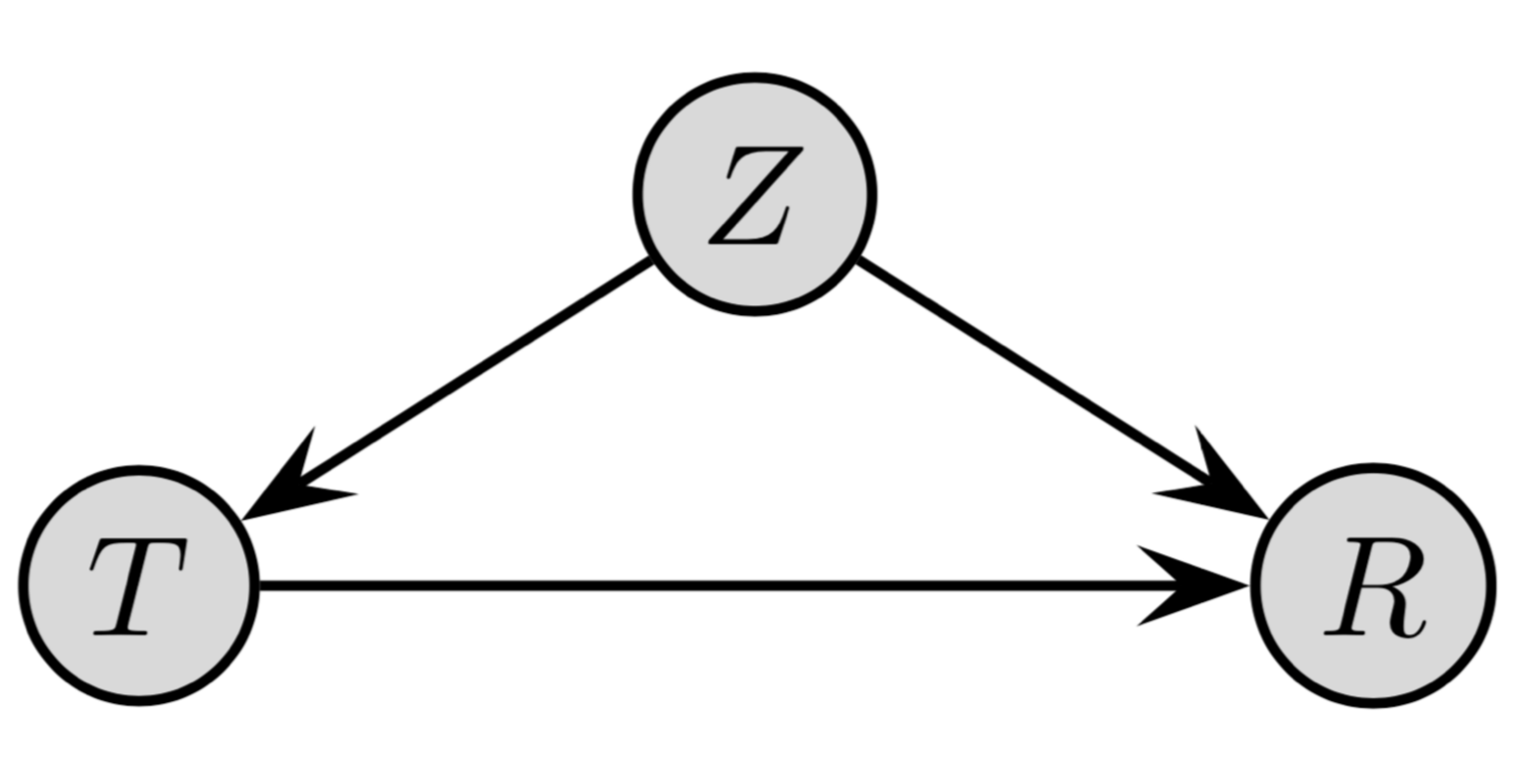}
\caption{Causal diagram for kidney stones.}\label{fig:kidney}
\end{figure}

\subsection{Confounding}
An intuitive explanation for this kidney stone example of Simpson's paradox is that larger stones are more severe than small stones and are much more likely to be treated with treatment $a$, resulting in that treatment $a$ looks worse than treatment $b$ overall. We assume that Figure \ref{fig:kidney} depicts the true underlying causal diagram of the kidney stone example. Confounding occurs because the size of kidney stones influences both treatment and recovery. The size of the kidney stones is called a confounder. The term ``confounding'' originally meant ``mixing'' in English, which describes that the true causal effect $T \rightarrow R$ is ``mixed'' with the spurious correlation between $T$ and $R$ induced by the fork $T \leftarrow Z \rightarrow R$ \citep{pearl2018bookofwhy}. In other words, we will not be able to disentangle the true effect of $T$ on $R$ from the spurious effect if we do not have data on $Z$. If we have measurements of $Z$ or can indirectly estimate $Z$, it is easy to deconfound the true and spurious effects. To this end, we can adjust for $Z$ by averaging the effect of $T$ on $R$ in each subgroup of $Z$ (i.e., different size groups in the case of kidney stones).

\subsection{Do-Operator and Adjustment Criterion} \label{sect:doopt}
From the viewpoint of causal inference, we can use the language of intervention, namely the \textit{do-operator}, to describe when confounding happens. In fact, in the example of the kidney stones, what we are interested in is how these two treatments compare when we force all patients to take treatment $a$ or $b$, rather than which treatment has a higher recovery rate given only the observational patient records. Mathematically, we focus on the true effect $p(R=1|do(T=a))$ (the \textit{intervention distribution} obtained when patients are forced to take treatment $a$) instead of the conditional $p(R=1|T=a)$ (the \textit{observational distribution} obtained when patients are observed to take treatment $a$). Therefore, as described previously, confounding can be naturally described as the discrepancy between $p(R|T)$ and $p(R|do(T))$.

The do-operator can be executed in two common ways: by Randomized Controlled Trials (RCTs) \citep{fisher1935design} and by adjustment formulas (i.e., Back-door criterion and Front-door criterion) \citep{pearl2009causality}. RCTs are the gold standard but sometimes limited by practical considerations (e.g., safety, laws, ethics, physically infeasibility, etc.). The Back-door and Front-door criteria require knowledge of the causal diagram, which typically means that causal assumptions are thus provided in advance. According to the Back-door criterion, in the kidney stone example, we can immediately obtain 
\begin{align} \label{eq:backdoor_equation}
p(R=1|do(T=a)) = \sum_{z\in \{l, s\}} p(R=1|T=a, Z = z)p(Z=z). 
\end{align}
Beyond the two adjustment formulas, the $do$-calculus provides a syntactic method of deriving claims about interventions \citep{pearl2009causality}. It consists of three rules which can be repeatedly applied to simplify the expression for an interventional distribution. With the do-calculus we can calculate intervention probabilities from observation probabilities.

\subsection{Proxy Variables for Confounding}

If confounders can be measured, then they can be adjusted for through the methods discussed in Section \ref{sect:doopt}. However, in most cases where confounders are hidden or unmeasured, without further assumptions, it is impossible to estimate the effect of the intervention on the outcome. A common practice is then to leverage observed proxy variables that contain information about the unobserved confounders \citep{angrist2008mostly,maddala1992introduction,montgomery2000measuring,Scholkopfetal2016}. However, using proxy variables to correctly recover causal effects requires strict mathematical assumptions \citep{louizos2017causal,edwards2015all,kuroki2014measurement,miao2018identifying,pearl2012measurement,wooldridge2009estimating} and, in practice, we do not know whether or not the available data meets those assumptions. Hence, we decide to follow \citet{louizos2017causal} and, instead of using proxy variables, we estimate a latent-variable model in which we simultaneously discover the latent confounders and infer how they affect treatment and outcome. 

\section{Deconfounding Reinforcement Learning}

In this section, we will formally introduce \textit{deconfounding reinforcement learning} (DRL). Generally speaking, DRL consists of two steps: learning a deconfounding model shown in Figure \ref{fig:deconRL} and optimizing a policy based on the learned deconfounding model. The main idea in step 1 is to simultaneously discover hidden confounders and infer causal effects through the estimation of a latent-variable model. More specifically, we first discuss the time-independent confounding assumption in Section \ref{sect:causalassump} and then, based on this assumption, we formalize the deconfounding model in Section \ref{sect:model}. Section \ref{sect:identification} talks about the problem of identification in our model, which is a central issue in causal inference. After that, we present details about how to learn the proposed model via variational inference in Section \ref{sect:learning}. Step 2 is provided in Section \ref{sect:DRL_algorithm} where we describe how to design the deconfounding actor-critic method. The proposed approach is straightforward to apply to other RL algorithms in the same manner. Finally, the training procedure 
for our deconfounding actor-critic method is presented in Section \ref{sect:training}.

\subsection{Causal Assumptions} \label{sect:causalassump}
\begin{figure}[t]
\centering
\includegraphics[width=0.55\linewidth]{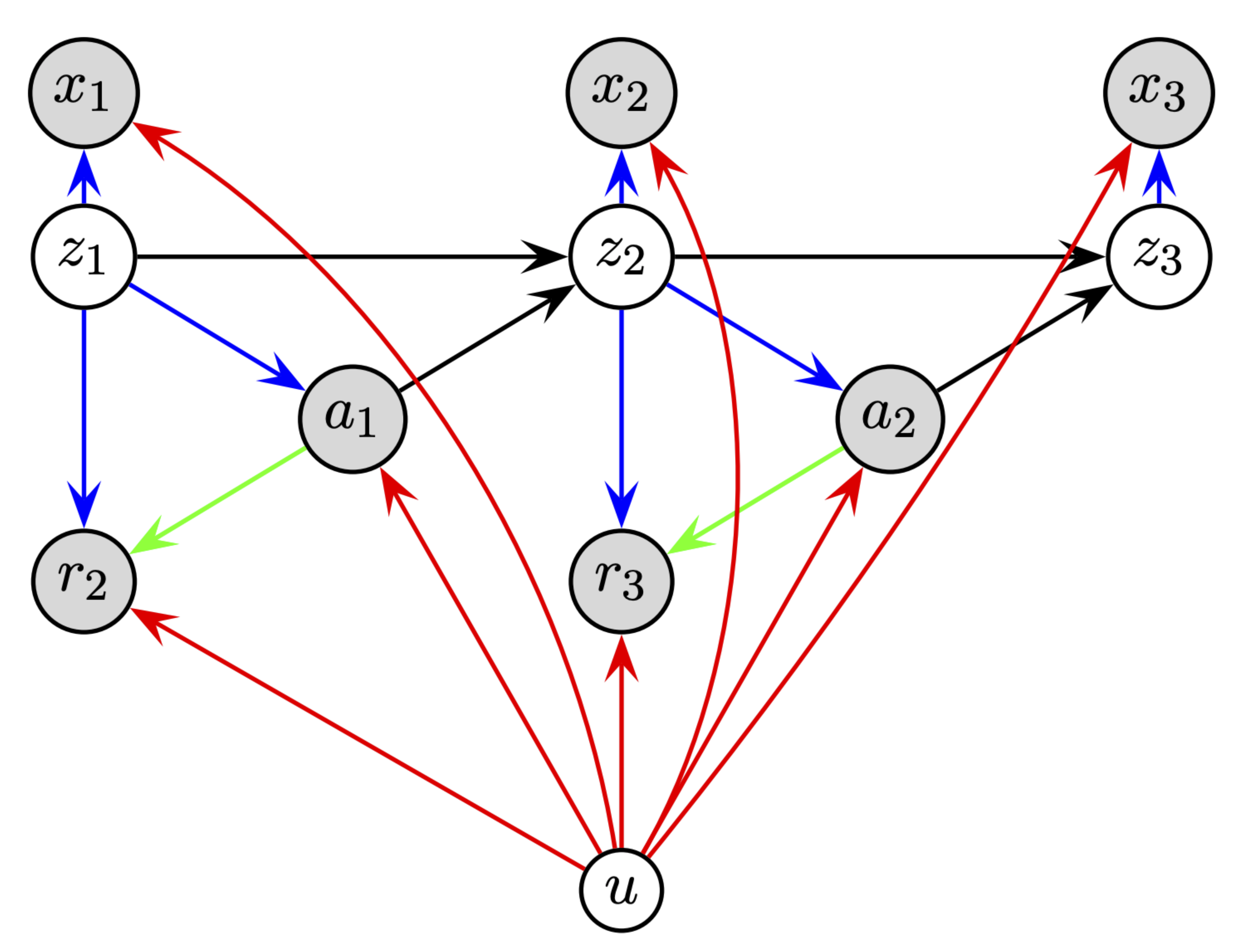}
\caption{The model for deconfounding reinforcement learning. Grey nodes denote observed variables and white nodes represent unobserved variables. \textcolor{red}{Red} and \textcolor{blue}{blue} arrows emphasize the observed variables affected by $u$ and by $z_t$, respectively. The causal effects of interest are colored in \textcolor{green}{green}.} \label{fig:deconRL} \vspace{-3mm}
\end{figure}

Without loss of generality, we assume that there exists a common confounder in the model, denoted by $u$ in Figure \ref{fig:deconRL}, which is time-independent across a number of episodes. This assumption is sufficiently general to apply to various RL tasks across domains. For example, in personalized medicine, socio-economic status can be a confounder affecting both the medication strategy a patient has access to and the patient's general health \citep{louizos2017causal}. In this case, socio-economic status is time-independent for each patient during the course of treatment. Another example is in agriculture, where soil fertility may serve as a time-independent confounder affecting both the application of fertilizer and the yield of each plot of land \citep{pearl2018bookofwhy}. Finally, in the stock market example, government policy may also act as a time-independent confounder. 

\subsection{The Model} \label{sect:model}
Given our causal assumptions, we first fit a generative model to a sequence of observational data: observations, actions, and rewards, where actions and rewards are confounded by one or several unknown factors, as shown in Figure \ref{fig:deconRL}. Formally, let $\vec{x}=(x_1, \ldots, x_T)$, $\vec{a}=(a_1, \ldots, a_{T-1})$, $\vec{r}=(r_2, \ldots, r_{T+1})$, $\vec{z}=(z_1, \ldots, z_T)$ be the sequences of observations, actions, rewards, and corresponding latent states, respectively. As mentioned previously, the confounder is denoted by $u$, and it is worth noting that here $u$ may stand for more than one confounder in which multiple confounders are seen as a whole represented by $u$ (i.e., the confounder $u$ can be a vector). We assume that $x_t \in \mathbb{R}^{D_x}$, $a_t \in \mathbb{R}^{D_a}$, $r_t \in \mathbb{R}^{D_r}$, $z_t \in \mathbb{R}^{D_z}$, and $u \in \mathbb{R}^{D_u}$, where $D_z \ll D_x$. The generative model for DRL is then given by
\begin{align}
&p(z_t) = \prod_{j=1}^{D_z}\mathcal{N}(z_{tj}|0, 1); \qquad 
p(u) = \prod_{j=1}^{D_u} \mathcal{N}(u_{j}|0, 1); \nonumber \\
&p(x_t | z_t, u) = \mathcal{N}\left(x_t | \hat{\mu_t^x}, \hat{\sigma_t^x}^2\right);
\quad \hat{\mu_t^x} = f_1(z_t,u), \quad \hat{\sigma_t^x}^2 = f_2(z_t,u); \label{eq:pxgz} \\
&p(a_t | z_t, u) = \mathcal{N}\left(a_t | \hat{\mu_t^a}, \hat{\sigma_t^a}^2\right);
\quad \hat{\mu_t^a} = f_3(z_t, u), \quad \hat{\sigma_t^a}^2 = f_4(z_t, u); \label{eq:pagzu}\\
&p(r_{t+1} | z_t, a_t, u) = \mathcal{N}\left(r_{t+1} | \hat{\mu_t^r}, \hat{\sigma_t^r}^2\right);
\quad \hat{\mu_t^r} = f_5(z_t, a_t, u), \quad \hat{\sigma_t^r}^2 = f_6(z_t, a_t, u); \\
&p(z_t | z_{t-1}, a_{t-1}) = \mathcal{N}\left(z_t | \hat{\mu_t^z}, \hat{\sigma_t^z}^2\right);
\quad \hat{\mu_t^z} = f_7(z_{t-1}, a_{t-1}), \quad \hat{\sigma_t^z}^2 = f_8(z_{t-1}, a_{t-1}); \label{eq:pzgza}
\end{align}
where we have parametrized each probability distribution as a Gaussian with mean and variance given by nonlinear functions $f_k$ represented by neural networks with parameters $\theta_k$ for $k=1, \ldots, 8$. Note that, in our case, Equation (\ref{eq:pagzu}) is not necessary when learning the model, because the data used in our experiments are generated from a policy depending only on the confounder (see Section \ref{sect:confoundingdata}), that is, $a_t$ does not depend on $z_t$ in our data. In addition, in this case $z_t$ is not viewed as a confounder of $a_t$ and $r_{t+1}$, for $z_t$ does not influence $a_t$. This also provides one reason why we do not need to adjust for $z_t$. However, in some other cases such as when applied to medical data, the strategies of treatment from physicians definitely contain valuable information about $z_t$ and $a_t$ (i.e., $z_t$ does influence $a_t$), and therefore the arrow between them is necessary when learning the model.

\subsection{Identification of Causal Effect} \label{sect:identification}
The key component of our method that allows us to address problems with confounders is the computation of the reward according to the model from Figure \ref{fig:deconRL}. To be more precise, assuming that an agent standing at state $z_t$ performs action $a_t=\mathfrak{a}$, we do not use the conditional $p(r_t | z_t, a_t=\mathfrak{a})$ as predictor for the reward as traditional RL methods do. Instead, we use the do-operator described in Section \ref{sect:doopt} to compute the reward\footnote{To give some intuition, take for example the case of kidney stones. One prefers treatment $b$ when considering the overall probability of recovery. By contrast, one chooses treatment $a$ when considering the recovery rate when conditioning on each possible kidney stone size. The optimal treatment is treatment $a$ in this example. The $do$-operator in Equation (\ref{eq:backdoor_equation}) allows us to obtain the correct solution in  this type of problems with confounders.  By contrast, traditional RL methods will fail in such settings since they consider only conditional probabilities (e.g., the overall probability of recovery) instead of interventional probabilities, as the $do$-operator does.}:
\begin{align} 
p(r_t | z_t, \text{do}(a_t=\mathfrak{a}))
&= \int_u p(r_t | z_t, \text{do}(a_t=\mathfrak{a}), u)p(u|z_t, \text{do}(a_t=\mathfrak{a})) \text{d}u \label{eq:deconreward_pre}\\
&=\int_u p(r_t | z_t, a_t=\mathfrak{a}, u) p(u) \text{d}u, \label{eq:deconreward}
\end{align}
where Equation (\ref{eq:deconreward}) is obtained by applying the rules of $do$-calculus to the causal graph in Figure \ref{fig:deconRL} \citep{pearl2009causality}. We also can use the back-door criterion to directly obtain Equation (\ref{eq:deconreward}). Equation (\ref{eq:deconreward}) shows that $p(r_t | z_t, \text{do}(a_t=\mathfrak{a}))$ can be identified from the joint distribution $p(u, \vec{z}, \vec{x}, \vec{a}, \vec{r})$, because, through Equation (\ref{eq:deconreward}), the interventional probability $p(r_t | z_t, \text{do}(a_t=\mathfrak{a}))$ is converted to observation probabilities with regard to $p(u, \vec{z}, \vec{x}, \vec{a}, \vec{r})$. In other words, if we can recover $p(u, \vec{z}, \vec{x}, \vec{a}, \vec{r})$ then we also can recover $p(r_t | z_t, \text{do}(a_t=\mathfrak{a}))$. 

Now the problem is reduced to whether or not we can estimate $p(u, \vec{z}, \vec{x}, \vec{a}, \vec{r})$ from observations of $(\vec{x}, \vec{a}, \vec{r})$. Fortunately, it is possible, because a number of works have shown that one can use the knowledge inferred from the joint distribution between proxy variables and confounders to adjust for the hidden confounders \citep{louizos2017causal,edwards2015all,kuroki2014measurement,miao2018identifying,pearl2012measurement,wooldridge2009estimating}. Here, we focus only on one possible case presented in Figure 1 of \citep{louizos2017causal} (see Appendix \ref{appendix:aproxyvariable}), because their result can be directly used to show that the joint distribution $p(u, \vec{z}, \vec{x}, \vec{a}, \vec{r})$ can be approximately recovered solely from the observations $(\vec{x}, \vec{a}, \vec{r})$. The main idea is that our model can be factorized into two types of 4-tuple components, each of which can be proved using the same method in \citep{louizos2017causal}. More precisely, in our model, at each time step the 4-tuple $(z_t, x_t, a_t, r_{t+1})$ is exactly the same as the 4-tuple $(Z, X, t, y)$ shown in Figure 1 of \citep{louizos2017causal}. Since it has been proved in \citep{louizos2017causal} that the joint distribution $p(Z, X, t, y)$ can be recovered from observations of $(X, t, y)$, $p(z_t, x_t, a_t, r_{t+1})$ can be also approximately recovered only from the observations $(x_t, a_t, r_{t+1})$. Likewise, the joint distribution over the other 4-tuple $(u, x_t, a_t, r_{t+1})$ can be recovered from $(x_t, a_t, r_{t+1})$ in the same manner (see Appendix \ref{appendix:analysis_deconm}). Applying this rule repeatedly to the sequential data, we are finally able to approximately recover $p(u, \vec{z}, \vec{x}, \vec{a}, \vec{r})$ solely from the observations $(\vec{x}, \vec{a}, \vec{r})$. Hence, in the present paper we may reasonably assume that the joint distribution $p(u, \vec{z}, \vec{x}, \vec{a}, \vec{r})$ can be approximately recovered solely from the observations $(\vec{x}, \vec{a}, \vec{r})$. 

We estimate the joint distribution $p(u, \vec{z}, \vec{x}, \vec{a}, \vec{r})$ using Variational Auto-Encoders (VAEs) \citep{kingma2013auto}, which can represent a very large class of distributions with latent variables and are easily implemented by solving an optimization problem \citep{tran2015variational,louizos2017causal}. However, VAEs do not guarantee that the true model can be identified through learning \citep{louizos2017causal}. 
For example, it might occur that different model parameters could equally fit the same observed data as long as the values of the latent confounders are adjusted accordingly \citep{Alexander2018identification}. Despite this, identification is possible in both our model and the one described by \citet{louizos2017causal} under specific conditions  \citep{louizos2017causal,kuroki2014measurement,miao2018identifying,pearl2012measurement,allman2009identifiability}. In addition to this, in our model, each of the $z_t$ and $u$ have at least $T$ proxy variables $\{x_t\}_{t=1,\ldots,T}$, where $z_t$ can be approximately viewed as hidden states in a hidden Markov model. In this case, \citet{allman2009identifiability} show that such a model can be identified if the latent confounders $z_t$ and $u$ are assumed to be categorical. However, in practice there is no guarantee for this to be the case. Hence, we prefer to use VAEs, which make substantially weaker assumptions about the data generating process and the structure of the hidden confounders \citep{louizos2017causal}. Despite the lack of general identifiability results for VAEs, we empirically show that our approach is useful for learning a better policy in the presence of confounding.  

In the model from Figure \ref{fig:deconRL} there are two types of confounders: the time-independent confounder $u$ and the time-dependent confounders $\{z_t\}_{t=1,\ldots,T}$, playing different roles in the model. We refer to the former as ``nuisance confounders'' and the latter as ``advantage confounders''. The nuisance confounder $u$ will negatively affect the whole course of treatment and, therefore, should be adjusted for. In the example of kidney stones, the existence of the confounder (i.e., the size of stones) will lead to a wrong treatment if not adjusted for. By contrast, the advantage confounders $\{z_t\}$ act as state variables and, in principle, they are not supposed to be adjusted for because in RL we aim to take optimal actions when conditioning on the current state\footnote{Another reason why we do not need to adjust for $z_t$ has been discussed in Section \ref{sect:model}.}. 

\subsection{Learning} \label{sect:learning}
The nonlinear functions parametrized by neural networks make inference intractable. Because of this, we learn the parameters of the model $\theta_k$ by using variational inference together with an inference model, a neural network which approximates the intractable posterior distributions \citep{rezende2014stochastic,kingma2013auto,krishnan2015deep}. We now review  how to learn a simple latent variable model $p_{\theta}(x|z)$ using variational inference.
In this simple model $x$ stands for the observational variables and $z$ for the latent variables.
By using the variational principle, we introduce an approximate posterior distribution $q_{\phi}(z|x)$ to obtain the following lower bound on the model's marginal likelihood:
\begin{align} \label{eq:vi}
\log p_{\theta}(x) \geq \underset{q_{\phi}(z|x)}{\mathbb{E}}\left[\log p_{\theta}(x | z)\right] - \text{KL}\left(q_{\phi}(z|x)||p_{\theta}(z)\right),
\end{align}
where we have used Jensen's inequality, and $\text{KL}(\cdot||\cdot)$ denotes the Kullback-Leibler divergence between two distributions
and $\phi$ are the parameters of the inference model $q(z|x)$. 

\subsubsection{Variational Lower Bound}
By directly applying the lower bound in Equation (\ref{eq:vi}) to our model, we obtain
\begin{align} \label{eq:mlb}
\log p_{\theta}(\vec{x}, \vec{a}, \vec{r}) 
\geq & \nonumber\\
\underset{q_{\phi}(\vec{z}, u|\vec{x}, \vec{a}, \vec{r})}{\mathbb{E}}&\left[\log p_{\theta}(\vec{x}, \vec{a}, \vec{r} , \vec{z}, u)\right] - \text{KL}\left(q_{\phi}(\vec{z}, u|\vec{x}, \vec{a}, \vec{r})||p_{\theta}(\vec{z}, u)\right) 
%\nonumber \\&
\equiv \mathcal{L}(\vec{x}, \vec{a}, \vec{r};\theta, \phi).
\end{align}
By using the Markov property of our model, we can factorize the full joint distribution in the following way:
\begin{align} \label{eq:mf}
p_{\theta}(\vec{x}, \vec{a}, \vec{r}, \vec{z}, u) = & p(u)p(z_1)\left[\prod_{t=1}^{T} p(x_t|z_t,u)p(a_t|z_t, u)p(r_{t+1}|z_t, a_t, u)\right]\nonumber\\
&\left[\prod_{t=2}^{T} p(z_t|z_{t-1}, a_{t-1})\right].
\end{align}
In addition, for simplicity, we also have a similar factorization assumption in the posterior approximation for $\vec{z}$ and $u$:
\begin{align} \label{eq:pf}
q_{\phi}(\vec{z}, u|\vec{x}, \vec{a}, \vec{r}) = q(u|\vec{x}, \vec{a}, \vec{r})q(z_1|\vec{x}, \vec{a}, \vec{r})\prod_{t=2}^{T} q(z_t|z_{t-1}, \vec{x}, \vec{a}, \vec{r}).
\end{align}
Combining equations (\ref{eq:mlb}), (\ref{eq:mf}) and (\ref{eq:pf}) yields
\begin{align} \label{eq:flb}
\mathcal{L}(\vec{x}, \vec{a}, \vec{r};\theta, \phi)
=& \sum_{t=1}^T \underset{\substack{z_t \sim q(z_t|z_{t-1}, \vec{x},\vec{a},\vec{r}) \\ u \sim q(u|\vec{x},\vec{a},\vec{r})}}{\mathbb{E}}
\left[\log p(x_t|z_t,u)+\log p(a_t|z_t, u)+\log p(r_{t+1}|z_t, a_t, u)\right] \nonumber\\
&-\text{KL}\left(q(u|\vec{x},\vec{a},\vec{r}) || p(u)\right) 
-\text{KL}\left(q(z_1|\vec{x},\vec{a},\vec{r}) || p(z_1)\right)\nonumber\\
&-\sum_{t=2}^T \underset{z_{t-1} \sim q(z_{t-1}|z_{t-2},\vec{x},\vec{a},\vec{r})}{\mathbb{E}} \left[\text{KL}\left(q(z_t|z_{t-1}, \vec{x},\vec{a},\vec{r}) || p(z_t|z_{t-1},a_{t-1})\right)\right],
\end{align}
where we omit the subscripts $\theta$ and $\phi$. A more detailed derivation can be found in Appendix \ref{appendix:vlb}. Equation (\ref{eq:flb}) is differentiable with respect to the model parameters $\theta$ and the inference parameters $\phi$ and, by using the reparametrization trick \citep{kingma2013auto}, we can directly apply backpropagation to optimize this objective. 

\subsubsection{Inference Model} \label{sect:infmodel}
From the factorization in Equation (\ref{eq:pf}), we can see that there are two types of inference models: $q(u|\vec{x}, \vec{a}, \vec{r})$ and $q(z_t|z_{t-1}, \vec{x}, \vec{a}, \vec{r})$. Similar to the generative model in Section \ref{sect:model}, we also parametrize both of them as Gaussian:
\begin{align}
&q(u|\vec{x}, \vec{a}, \vec{r}) = \mathcal{N}\left(u | \hat{\mu_t^u}, \hat{\sigma_t^u}^2\right);
\quad \hat{\mu_t^u} = f_9(\vec{x}, \vec{a}, \vec{r}), \quad \hat{\sigma_t^u}^2 = f_{10}(\vec{x}, \vec{a}, \vec{r}); \label{eq:qu}\\
&q(z_t|z_{t-1}, \vec{x}, \vec{a}, \vec{r}) = \mathcal{N}\left(z_t | \hat{\mu_t^{\mathfrak{z}}}, \hat{\sigma_t^{\mathfrak{z}}}^2\right);
\quad \hat{\mu_t^{\mathfrak{z}}} = f_{11}(z_{t-1}, \vec{x}, \vec{a}, \vec{r}), \quad \hat{\sigma_t^{\mathfrak{z}}}^2 = f_{12}(z_{t-1}, \vec{x}, \vec{a}, \vec{r}). \label{eq:qz}
\end{align}
In fact, as shown in Equation (\ref{eq:pf}), $q(\vec{z}|\vec{x}, \vec{a}, \vec{r})$ can be factorized as the product of $q(z_t|z_{t-1}, \vec{x}, \vec{a}, \vec{r})$ for $t=1, \ldots, T$. By using the Markov property of our model, we have
\begin{align}
    z_t \indep x_1,\ldots, x_{t-1}, a_1, \ldots, a_{t-2}, r_2, \ldots, r_{t} | z_{t-1}.
\end{align}
Therefore, we can further simplify each of these terms as follows:
\begin{align} \label{eq:qsim}
q(z_t|z_{t-1}, \vec{x},\vec{a},\vec{r})=q(z_t|\underbrace{z_{t-1},a_{t-1}}_{\text{past}}, \underbrace{x_t,a_t,r_{t+1}}_{\text{present}},\underbrace{x_{t+1},a_{t+1},r_{t+2},\ldots,x_T,a_T,r_{T+1}}_{\text{future}}).
\end{align}
Equation (\ref{eq:qsim}) tells us that $z_t$ depends on $z_{t-1}$ and all the current and future observed data in $(\vec{x},\vec{a},\vec{r})$. Meanwhile, the conditional independence above means that $z_{t-1}$ contains all the historical data. Therefore, it is natural to calculate $z_t$ based on the whole sequence of data. This can be done using recurrent neural networks (RNNs). Inspired by \citet{krishnan2015deep,krishnan2017structured}, we similarly choose a bi-directional LSTM \citep{zaremba2014learning} to parameterize $f_{11}$ and $f_{12}$ in Equation (\ref{eq:qz}). Since Equation (\ref{eq:qu}) has the same structure as Equation (\ref{eq:qz}), $f_{9}$ and $f_{10}$ are parameterized by a bi-directional LSTM as well. More details about the architecture can be found in Appendix \ref{appendix:lstm}.

Note that to generate data from the model, at any time step $t$, we require to know $a_t$ and $r_{t+1}$ before inferring the distribution over $z_t$ when conditioning on the observed $x_t$. Hence, we need to introduce two auxiliary distributions to perform counterfactual reasoning. This corresponds to the problem of predicting $x_{t+1}$ given an $x_t$ unseen in the training set. To be more precise, we have
\begin{align}
&q(a_t|x_t) = \mathcal{N}\left(a_t|\hat{\mu_t^{\mathfrak{a}}}, \hat{\sigma_t^{\mathfrak{a}}}^2\right);
\quad \hat{\mu_t^{\mathfrak{a}}} = f_{13}(x_t), \quad \hat{\sigma_t^{\mathfrak{a}}}^2 = f_{14}(x_t); \label{eq:qagx}\\
&q(r_{t+1}|x_t, a_t) = \mathcal{N}\left(r_{t+1}|\hat{\mu_t^{\mathfrak{r}}}, \hat{\sigma_t^{\mathfrak{r}}}^2\right);
\quad \hat{\mu_t^{\mathfrak{r}}} = f_{15}(x_t, a_t), \quad \hat{\sigma_t^{\mathfrak{r}}}^2 = f_{16}(x_t, a_t),\label{eq:qrgxa}
\end{align}
where $f_{13}$, $f_{14}$, $f_{15}$, and $f_{16}$ are also parameterized by neural networks. To estimate the parameters of these inference models, we will add extra terms in the variational lower bound given by Equation (\ref{eq:flb}):
\begin{align} \label{eq:drl}
    \mathcal{L}_{\text{DRL}}(\vec{x}, \vec{a}, \vec{r};\theta, \phi) = \mathcal{L}(\vec{x}, \vec{a}, \vec{r};\theta, \phi) + \sum_{t=1}^T\left(\log q(a_t|x_t) + \log q(r_{t+1}|x_t, a_t)\right).
\end{align}

\subsection{Deconfounding RL Algorithms} \label{sect:DRL_algorithm}
We have now all the building blocks for our DRL algorithm. Once our model is learned from the observational data, it can be directly used as a dynamic environment like those in OpenAI Gym \citep{openaigym}. We can exploit the learned model to generate rollouts for policy evaluation and optimization. In practice, Equation (\ref{eq:deconreward}) is approximated using the Monte Carlo method as follows:
\begin{align} \label{eq:deconrl}
p(r_t | z_t=\mathfrak{z}, \text{do}(a_t=\mathfrak{a}))=\frac{1}{N}\sum_{i=1}^N p(r_t | z_t=\mathfrak{z}, a_t=\mathfrak{a}, u_i) \quad
u_i \sim p(u),
\end{align}
where $N$ is the sample size from the prior $p(u)$. Given observational data, we sample $u$ instead from the approximate posterior $q(u|\vec{x}, \vec{a}, \vec{r})$ which we compute using the inference network presented in Section \ref{sect:infmodel}.

By using this deconfounding reward function, it is straightforward to extend traditional RL algorithms to their corresponding deconfounding version. In this paper, we select and extend one representative RL algorithm: the Actor-Critic method \citep{sutton1998reinforcement}. Nevertheless, our methodology can be used to extend other algorithms as well in a straightforward manner.

\paragraph{Deconfounding Actor-Critic Methods} The actor-critic method is a policy-based RL method directly parameterizing a policy function $\pi(a|z; \theta)$. The goal is to reduce variance in the estimate of the policy gradient by subtracting a learned function of the state, known as a baseline, from the return. The learned value function $V(z; \phi)$ is commonly used as the baseline. Taking into consideration that the return is estimated by $Q(z, a; \phi_Q)$, we can write the gradient of the actor-critic loss function at step time $t$ as
\begin{align} \label{eq:ac}
\nabla J(\theta) = \mathbb{E}_{\pi}\left[\left(Q(z_t, a_t; \phi_Q)-V(z_t; \phi_V)\right)\nabla_{\theta}\ln\pi(a_t|z_t; \theta)\right],
\end{align}
where $Q(z_t, a_t; \phi_Q)-V(z_t; \phi_V)$ is an estimate of the advantage of action $a_t$ in state $z_t$. In practice, $Q(z_t, a_t; \phi_Q)$ is usually replaced with the one-step return, that is, $r_{t+1}+V(z_{t+1}; \phi_V)$. The crucial difference in deconfounding actor-critic methods is to use $r_{t+1} \sim p(r_{t+1}|z_t, \text{do}(a_t))$ given by Equation (\ref{eq:deconrl}), as opposed to the $r_{t+1} \sim p(r_{t+1}|z_t, a_t)$ used by vanilla actor-critic methods.

\subsection{Training} \label{sect:training}

As mentioned previously, DRL consists of two steps: learning a deconfounding model and optimizing a policy based on the learned deconfounding model. In step 1 we learn the model by optimizing the objective
given by Equation (\ref{eq:drl}). Once the deconfounding model is learned, we have an estimate of the state transition function $p(z_t | z_{t-1}, a_{t-1})$, as given by the model, and can also calculate the deconfounding reward function $p(r_t | z_t, do(a_t))$ according to Equation (\ref{eq:deconrl}). In step 2, we treat the learned deconfounding model as an RL environment like CartPole in OpenAI Gym and generate trajectories/rollouts using the estimated state transition function and deconfounding reward function. These trajectories/rollouts are then used to train the policy by following the gradient given by Equation (\ref{eq:ac}).

\subsection{Implementation Details} \label{sect:imp_details}

We used Tensorflow \citep{abadi2016tensorflow} for the implementation of our model and the proposed DRL algorithm. Optimization was done using Adam \citep{kingma2014adam}. Unless stated otherwise, the setting of all hyperparameters and network architectures can be found in Appendix \ref{appendix:setting}. 

To assess the equality of the learned model, we performed two types of tasks: reconstruction and counterfactual reasoning. The reconstructions were performed by first feeding an input sequence $\vec{x}$ into the learned inference network, then sampling from the resulting posterior distribution over $\vec{z}$ according to Equation (\ref{eq:qz}), and finally feeding those samples into the generative network described in Equation (\ref{eq:pxgz}) to reconstruct the original observed sequence $\vec{x}$. The counterfactual reasoning, that is, the prediction of $x_{t+1}$ given an $x_t$ that we have not seen during training, was executed in four steps:
\begin{enumerate}
    \item[1)] Given the new $x_t$, we estimate $a_t$ and $r_{t+1}$ using equations (\ref{eq:qagx}) and (\ref{eq:qrgxa}). \item[2)] Once we have $x_t$, $a_t$, and $r_{t+1}$, we estimate $z_t$ using Equation (\ref{eq:qz}). 
    \item[3)] Given the estimated $z_t$ and another uniformly randomly selected $a_t'$, we can directly compute $z_{t+1}$ from Equation (\ref{eq:pzgza}). 
    \item[4)] The final step is to reconstruct $x_{t+1}$ from $z_{t+1}$ and $u$ according to Equation (\ref{eq:pxgz}). 
\end{enumerate}
By repeating the last two steps, we can counterfactually reason out a sequence of data. 

We may also be interested in estimating the approximate posterior over the confounder $u$ from observed data. We consider two possible scenarios for this. In the easy one we are given the observed data $(\vec{x}, \vec{a}, \vec{r})$ and we estimate the posterior by using Equation (\ref{eq:qu}). The more challenging scenario involves estimating the posterior from only $\vec{x}$ and no action or reward data. In this case, we follow the same steps used in the task of counterfactual reasoning. We first compute $a_t$ and $r_{t+1}$ for $t=1$ to $T$ using Equations (\ref{eq:qagx}) and (\ref{eq:qrgxa}), and then estimate the posterior through Equation (\ref{eq:qu}). 

\section{Experimental Results}

The evaluation of any method that deals with confounders is always challenging due to a lack of real-world benchmark problems with known ground truth. Furthermore, little work has been done before on the task of deconfounding RL. All this creates difficulties in the evaluation of our algorithms and motivates us to develop several new benchmark problems by revising the MNIST dataset \citep{lecun1998gradient} and by revising two environments in OpenAI Gym \citep{openaigym}, CartPole and Pendulum.

\subsection{New Confounding Benchmarks} \label{sect:confoundingdata}

We now describe three new confounding benchmark problems, all of them including a single binary confounder.
The procedure to create these benchmarks is inspired by \citet{krishnan2015deep}, who synthesized a dataset mimicking healthcare data under realistic conditions (e.g., noisy laboratory measurements, surgeries and drugs affected by patient age and gender, etc.). The data
used is from either popular RL environments in OpenAI Gym such as Pendulum and CartPole or popular machine learning datasets such as MNIST.

\paragraph{Confounding Pendulum and CartPole}
We revised two environments in OpenAI Gym \citep{openaigym}: Pendulum and CartPole. To obtain a confounding version of Pendulum, we selected $100$ different screen images of Pendulum and created a synthetic problem in which actions are joint effort\footnote{More details can be found at \url{https://github.com/openai/gym/wiki/Pendulum-v0}.} with values between $-2$ and $2$. We added $20\%$ bit-flip noise to each image and then, a random policy confounded with a binary factor was used to select the action applied at each time step. This is repeated multiple times to produce a large number of 5-step sequences of noisy images. In each generated sequence, one block with three consecutive squares ($2 \times 2$ pixels) is superimposed with the top-left corner of the images in a random starting location. These squares are intended to be analogous to seasonal flu or other ailments that a patient could exhibit, which are independent of the actions taken and which last several time steps \citep{krishnan2015deep}. The goal is to show that our model can learn long-range patterns since these play an important role in medical applications. We treat the generated sequences of images as the observations $\vec{x}$. The training, validation and test sets respectively comprise $140,000$, $28,000$ and $28,000$ sequences of length five.

The key characteristic of confounding Pendulum is the relationship between confounder, actions, and rewards. For simplicity of notation, we denote the confounder by $u$, the action by $a$, and the reward by $r$. In this case, $u$ is a binary variable mimicking socio-economic status (i.e., the rich and the poor). The range of actions $a \in [-2,2]$ is grouped into two categories $T_1$ (i.e., $1 \leq |a| \leq 2$) and $T_2$ (i.e., $0 \leq |a| \leq 1$) representing different treatments\footnote{Note that $T_1$ is a better treatment than $T_2$ in our setting described in Appendix \ref{appendix:confoundingdatasets}.}. The treatment selection depends on $u$, that is, $u$ determines the probabilities of choosing $T_1$ and $T_2$. The reward is defined as
\begin{align}
    r = r_{\text{o}} + r_{\text{c}}, 
\end{align}
where $r_o$ is the original reward\footnote{In fact, here $r_o$ is a function of both $a$ and a state, and we mention only $a$ to emphasize that the confounder affects the action.} in Pendulum as a function of $a$, and $r_c$ is the extra reward as a function of both $u$ and $a$, defined by a two-component Gaussian mixture: 
\begin{align}
   & r_c \sim \sum\nolimits_{r_c\in\{R_1, R_2\}}\pi(r_c|a,u)\mathcal{N}(\mu_{r_c}, \sigma^2),
\end{align}
with $\sigma$ and $\mu_{r_c}$ fixed and mixing coefficient $\pi(r_c|a,u)$ determined by both $u$ and $a$. More details are available in Appendix \ref{appendix:confoundingdatasets}. 

Obviously, in the definition above, $r$ depends on $a$ and $u$. Furthermore, $u$ has an influence on $x_t$ through $a_{t-1}$ and $z_t$, meaning that $x_t$ contains information about $u$ and, therefore, can be viewed as a proxy variable for the confounder. We assume that the influence between the confounder, action and reward is stochastic while generating the data. The reason for this is that, in practice, we do not have an oracle that tells us which treatment is better so we might make wrong decisions sometimes. For example, take the case of kidney stones. Even though treatment $a$ is better than treatment $b$, there are still some patients choosing treatment $b$ with positive probability in each category.
%, because no oracle tells them what is the correct choice. 
All the details about the data generation process can be found in Appendix \ref{appendix:confoundingdatasets}, where a straightforward analogy is provided as well. 

The confounding CartPole problem is implemented in the exact same manner, except for the action which is now binary and can be naturally divided into two categories\footnote{More details can be found at https://github.com/openai/gym/wiki/CartPole-v0.}. Accordingly, the binary confounder determines the probabilities of choosing which of two actions, and the probabilities are set to the same value as those in the confounding Pendulum given in Appendix \ref{appendix:confoundingdatasets}.  

\paragraph{Confounding MNIST} We follow the same procedure to obtain a confounding MNIST problem. However, the definitions of action and the original reward term are now different. Similar to the Healing MNIST dataset \citep{krishnan2015deep}, the actions encode rotations of digit images with each entry in $\vec{a}$ satisfying $(-\frac{\pi}{4} \leq a \leq +\frac{\pi}{4})$ and the actions being divided into two categories $(\frac{\pi}{8} \leq |a| \leq \frac{\pi}{4})$ and $(0\degree \leq |a| < \frac{\pi}{8})$ according to the confounder $u$. The original reward term $r_o$ is defined as the minus degree between the upright position and the position that the digit rotates to. For example, if the digit rotates to the position of 3 o'clock or 9 o'clock, then both rewards are $-\frac{\pi}{2}$. 

\begin{figure}[t]
\centering
\includegraphics[width=\linewidth]{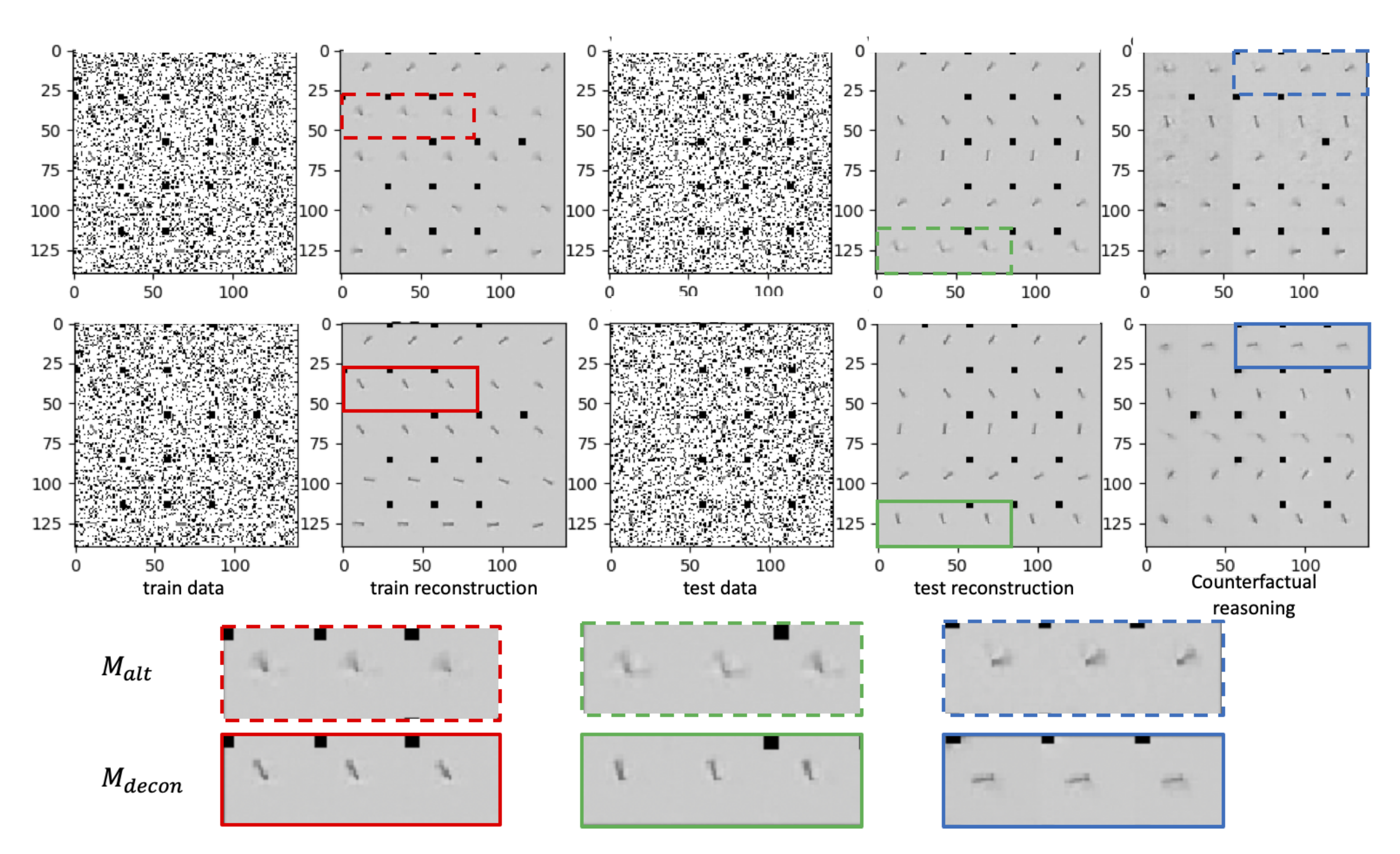}
\caption{Reconstruction and counterfactual reasoning on the confounding Pendulum dataset. Top row: results from the model without the confounder $u$ ($M_{\text{alt}}$). Second row: results from the model with the confounder $u$ ($M_{\text{decon}}$). The last two rows are the zoom of samples selected in the same positions from the top row (dashed boxes for $M_{\text{alt}}$) and the second row (solid boxes for $M_{\text{decon}}$), respectively. These results show that the model taking into account confounding performs better in every task producing less blurry images.} \label{fig:pendulum_demo}
\end{figure}
\begin{figure}[h]
\centering
\includegraphics[width=\linewidth]{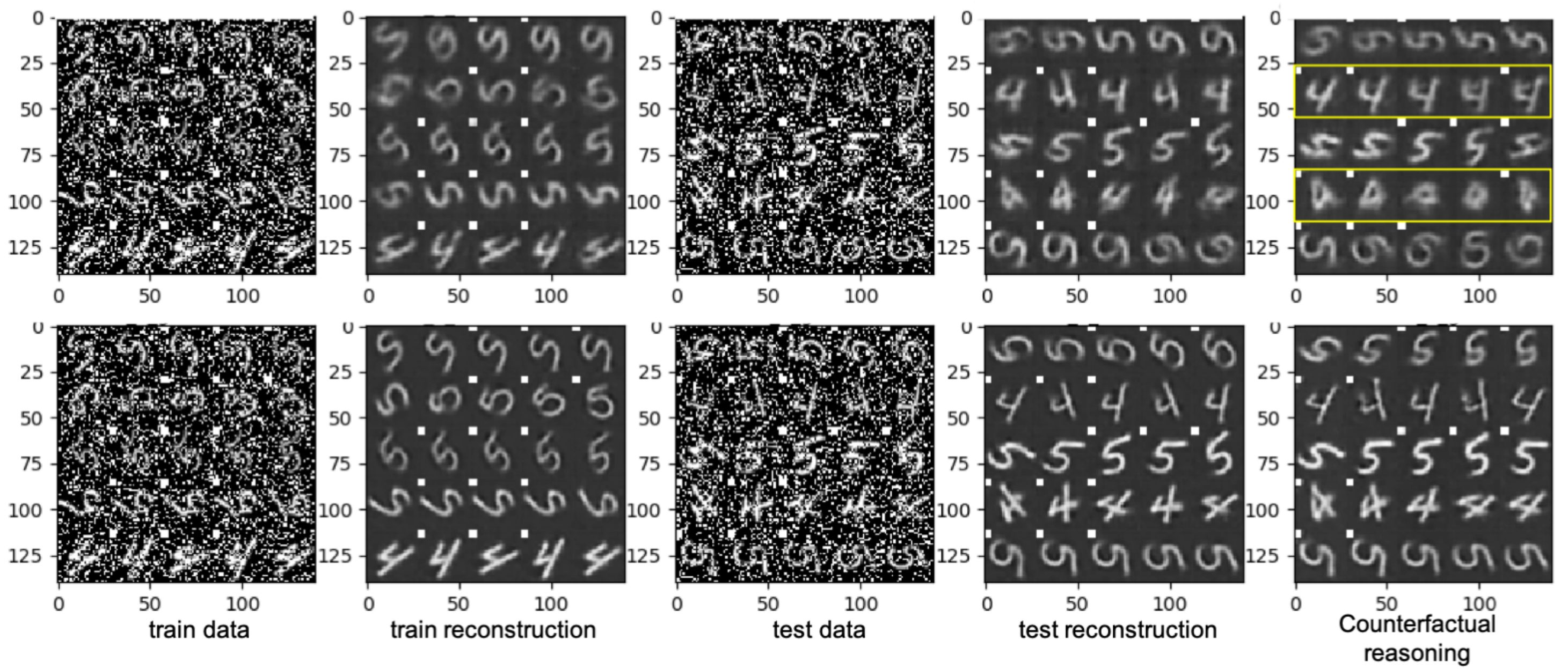}
\caption{Reconstruction and counterfactual reasoning on the confounding MNIST dataset. Top row: results from the model without the confounder $u$ ($M_{\text{alt}}$). Bottom row: results from the model with the confounder $u$ ($M_{\text{decon}}$). Note that in the task of counterfactual reasoning, the second and forth sequences of the sample produced by $M_{\text{alt}}$, boxed in yellow, have non-consecutive white squares and contain more blurry images. This does not really make sense because only consecutive patterns appear in the training set. By contrast, this kind of non-consecutive white squares does not occur on the samples from $M_{\text{decon}}$.} \label{fig:mnist_demo}
\end{figure}

\begin{figure}[h]
\centering
\raisebox{23mm}{(a)}
\includegraphics[width=0.45\linewidth]{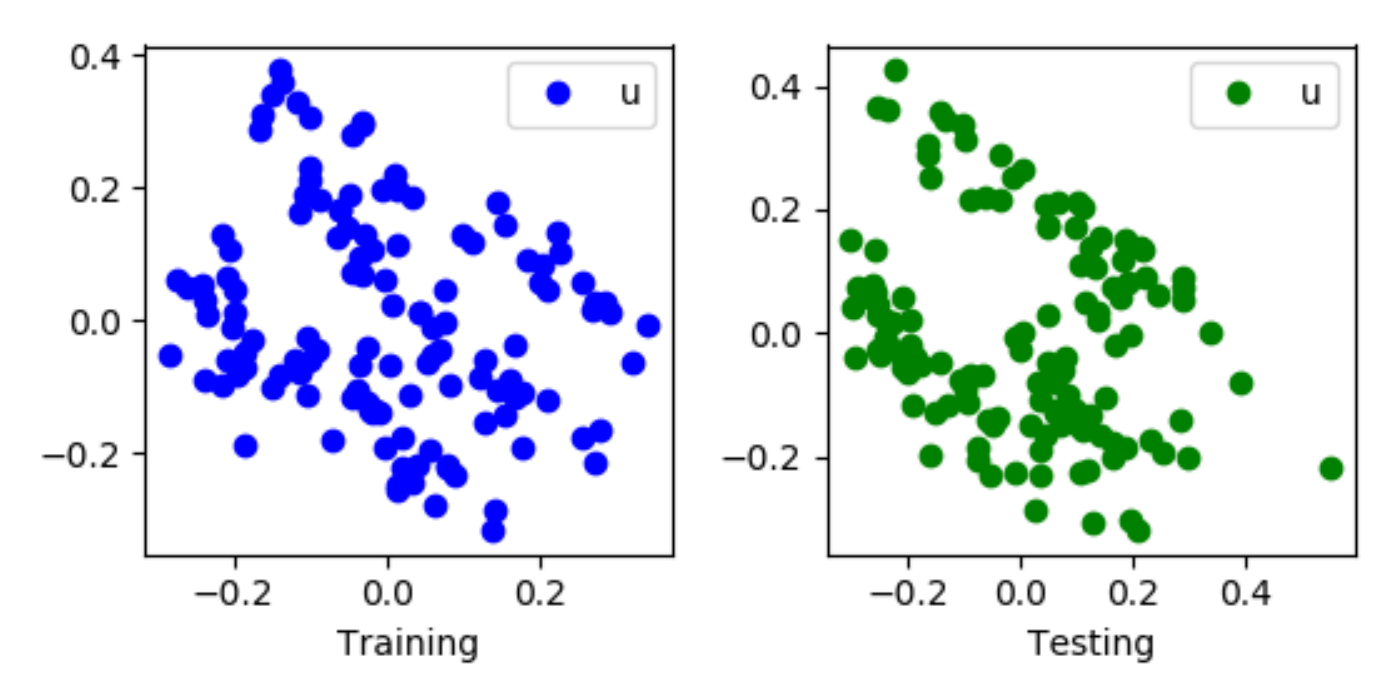}
\raisebox{23mm}{(b)}
\includegraphics[width=0.45\linewidth]{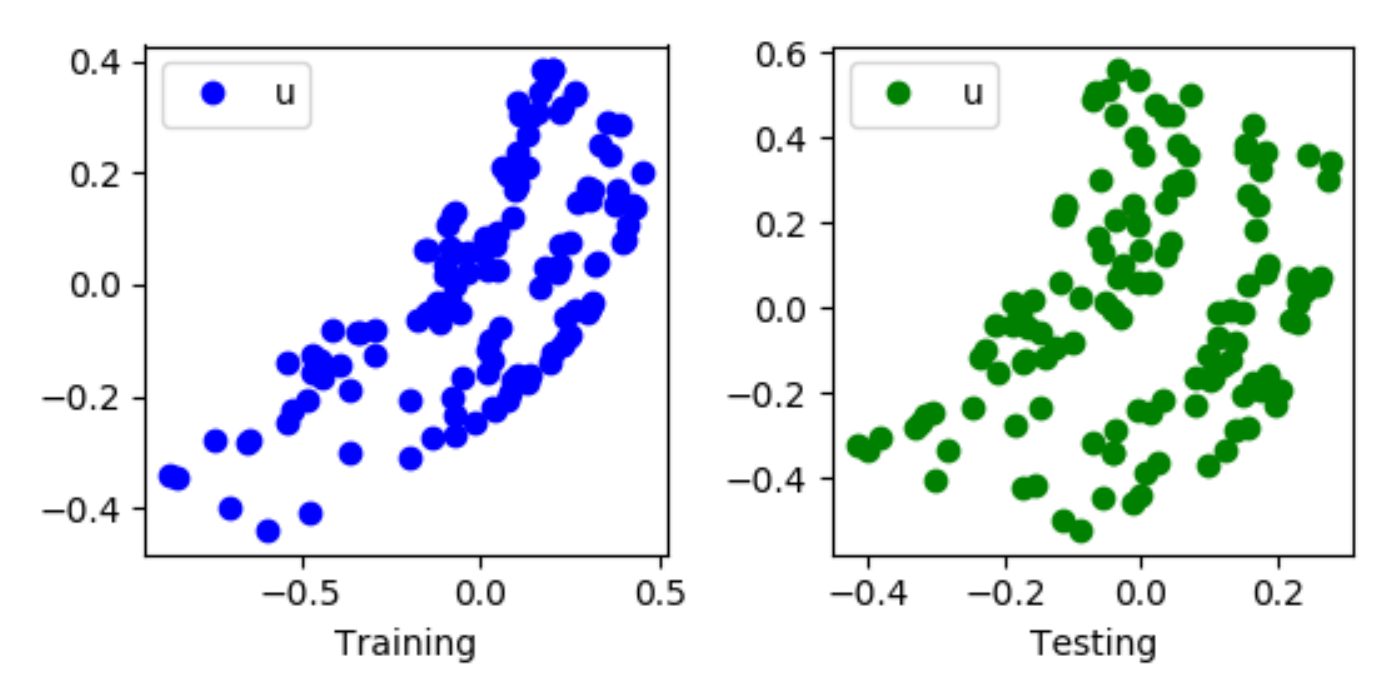}
\caption{(a) Plot of $128$ data points sampled from the posterior approximate distribution of $u$ on the confounding MNIST dataset. (b) Plot of $128$ data points sampled from the posterior approximate distribution of $u$ on the confounding CartPole dataset. We can see that, given a binary confounder $u$, our model still identifies two clear clusters from the data even if the assumed prior over $u$ is a factorized standard Gaussian distribution.} \label{fig:u_plot}
\end{figure}

\subsection{Performance Analysis of the Deconfounding Model}

We assess the performance of the deconfounding model from Figure \ref{fig:deconRL}, denoted by $M_{\text{decon}}$, and compare it with a similar alternative model that does not include the confounder $u$ and that is denoted by $M_{\text{alt}}$. We train $M_{\text{decon}}$ by optimizing Equation (\ref{eq:drl}) but train $M_{\text{alt}}$ using a different loss function which excludes the confounder $u$ and whose full derivation can be found in Appendix \ref{appendix:vlb_orin}. Both models are separately trained in a minibatch manner on the training set of the confounding dataset (i.e., $140K$ image sequences of length five). Afterwards, following the steps depicted in Section \ref{sect:imp_details}, we use each trained model to perform the reconstruction task on the training set, and both reconstruction and counterfactual reasoning tasks on the testing set (i.e., $28K$ image sequences of length five).  

Figure \ref{fig:pendulum_demo} presents a comparison of $M_{\text{decon}}$ and $M_{\text{alt}}$ in terms of reconstruction and counterfactual reasoning on confounding Pendulum. The second row is based on $M_{\text{decon}}$ whilst the top row comes from $M_{\text{alt}}$. The results generated by the deconfounding model are superior to those produced by the model not taking into account the confounder. To be more specific, as shown in the zoom of samples on the bottom row, $M_{\text{alt}}$ generates more blurry images than $M_{\text{decon}}$ because, without modelling the confounder $u$, $M_{\text{alt}}$ is forced to average over its multiple latent states, resulting in more blurry samples. 

We obtain similar results regarding the samples produced by $M_{\text{alt}}$ and $M_{\text{decon}}$ on the confounding MNIST dataset, as shown in Figure \ref{fig:mnist_demo}. Looking closely at the squares on the generated digit samples (inside a yellow box), we observe that $M_{\text{alt}}$ generates non-consecutive white squares in the task of counterfactual reasoning, which does not really make sense because only consecutive patterns appear in the training set. The generated images are also more blurry. By contrast, this does not occur on the samples from $M_{\text{decon}}$, showing that our deconfounding model is able to cope with long-range patterns and describe the data better. 

Figure \ref{fig:u_plot} visualizes approximate posterior samples of the $2$-dimensional confounder $u$. We can see that, although the prior distribution of the confounder is assumed to be a factorized standard Gaussian distribution, the model still identifies two clear clusters from the data because $u$ is originally a binary variable. This demonstrates that our model can learn confounders even if the assumed prior is not that accurate. 

\subsection{Comparison of RL Algorithms} \label{sect:comparerewards}

In this section, we evaluate the proposed deconfounding actor-critic (AC) method by comparing with its vanilla version on confounding Pendulum. In the vanilla AC method, given a learned $M_{\text{alt}}$, we optimize the policy by calculating the gradient presented in Equation (\ref{eq:ac}) on the basis of the trajectories/rollouts generated through $M_{\text{alt}}$. Equation (\ref{eq:ac}) involves two functions: $V(z_t;\phi_V)$ and $\pi(a_t | z_t; \theta)$, both of which are represented using neural networks and their corresponding hyperparameters can be found in Appendix \ref{appendix:setting}. It is worth noting that, in this vanilla case, each reward $r_{t+1}$ is sampled from the conditional distribution $p(r_{t+1} | z_t, a_t)$. By contrast, the proposed deconfounding AC method uses $M_{\text{decon}}$ instead. The optimizaiton of the policy is again performed using the gradient from Equation (\ref{eq:ac}). However, for the deconfounding AC approach, we use trajectories/rollouts generated by $M_{\text{decon}}$ in which each reward $r_{t+1}$ is obtained using the interventional distribution $p(r_{t+1} | z_t, \text{do}(a_t))$ computed according to Equation (\ref{eq:deconrl}). For completeness, we also compare with the direct AC method in which the AC method is directly trained on the training data instead of the trajectories/rollouts. 

\begin{figure}[t]
\centering
\includegraphics[width=\linewidth]{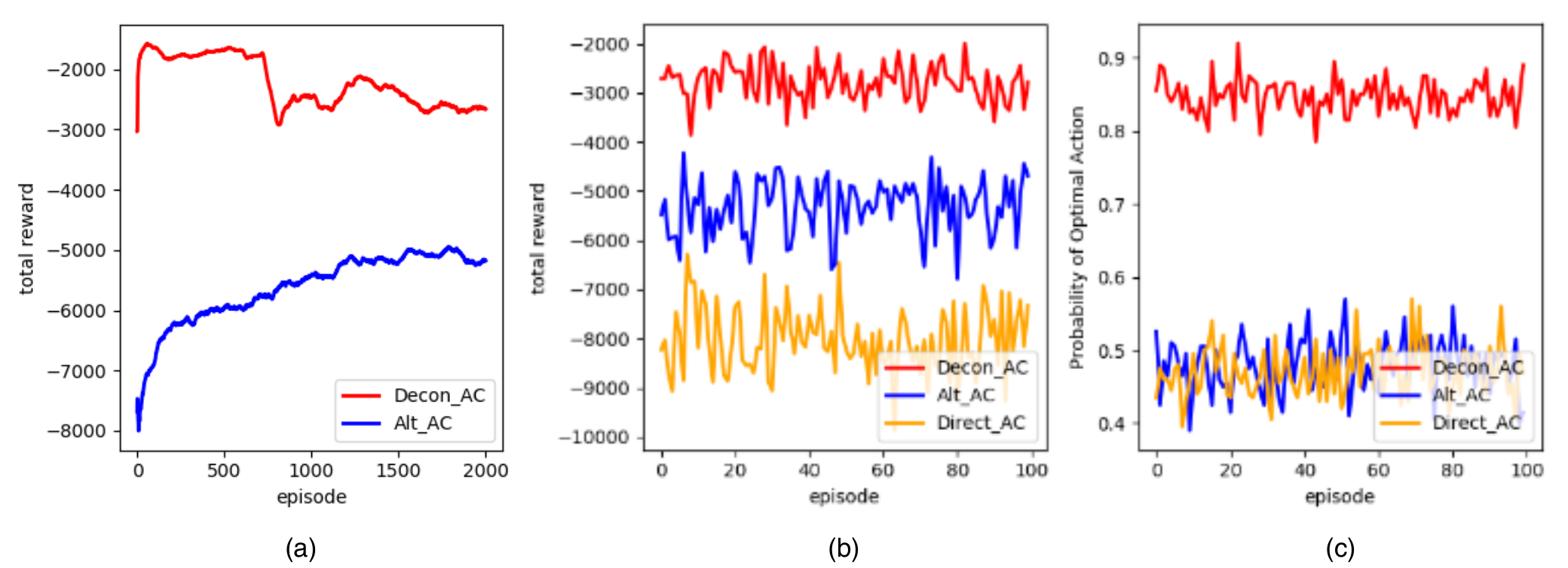}
\caption{Comparison of vanilla (Alt\_AC), direct (Direct\_AC), and deconfounding (Decon\_AC) Actor-Critic methods on the confounding Pendulum problem. (a) total reward over $1500$ episodes in the training phase. (b) total reward over $100$ episodes in the testing phase. (c) Probability of optimal action over $100$ episodes, corresponding to (b).} \label{fig:reward}
\end{figure}

In the training phase, we respectively run the vanilla AC and the deconfounding AC algorithms over $2000$ episodes with $200$ time steps each. For the fair comparison, we also run the direct AC algorithm over $2000$ episodes, each of which has $200$ pairs of state-action-reward randomly selected from the training data. In order to reduce non-stationarity and to decorrelate updates, the generated data is stored in an experience replay memory and then randomly sampled in a batch manner \citep{mnih2013playing,riedmiller2005neural,schulman2015trust,van2016deep}. We summarize all the rewards produced during the rollouts in each episode and further average the summarized rewards over a window of $100$ episodes to obtain a smoother curve. Figure \ref{fig:reward}(a) shows that
our deconfounding AC algorithm performs significantly better than the vanilla AC algorithm in the confounded environment. Here, the direct AC algorithm is not included because it does not generate rollouts.

In the testing phase, we first randomly select $100$ samples from the testing set, each starting a new episode, and then use the learned policies to generate trajectories over $200$ time steps as we did during training. From the resulting $100$ episodes, we plot the total reward obtained by each method, see in Figure \ref{fig:reward}(b), and also plot the percentage of times that the optimal action $T_1$ is chosen in each episode, as shown in Figure \ref{fig:reward}(c). Figure \ref{fig:reward}(b) shows that our deconfounding AC obtains on average much higher reward at test time than the baselines. Figure \ref{fig:reward}(c) further tells us that our deconfounding AC is much more likely to choose the optimal action at each time step, whilst the vanilla AC and the direct AC make a wrong decision more than half of the times.

\section{Related Work}
\cite{krishnan2015deep,krishnan2017structured} used deep neural networks to construct nonlinear state space models and leveraged a structured variational approximation parameterized by recurrent neural networks to mimic the posterior distribution. \cite{levine2018reinforcement} reformulated RL and control problems using probabilistic inference, which allows us to bring to bear a large pool of approximate inference methods, and flexibly extend the models. \cite{raghu2017deep,raghu2017continuous} exploited continuous state-space models and deep RL to obtain improved treatment policies for septic patients from observational data. \cite{gottesman2018evaluating} discussed problems when evaluating RL algorithms in observational health setting. However, all the works mentioned above do not take into account confounders in the proposed models.

\cite{louizos2017causal} attempted to learn individual-level causal effects from observational data in the non-temporal setting. They also used a variational auto-encoder to estimate the unknown confounder given a causal graph. \cite{paxton2013developing} developed predictive models based on electronic medical records without using causal inference. \cite{saria2010learning} proposed a nonparametric Bayesian method to analyze clinical temporal data. \cite{soleimani2017treatment} represented the treatment response curves using linear time-invariant dynamical systems. This provides a flexible approach to modeling response over time. Although the latter two works model sequential data, they both do not consider RL or causal inference.

\cite{bareinboim2015bandits} considered the problem of bandits with unobserved confounders. \cite{sen2016contextual} and \cite{ramoly2017causal} further studied contextual bandits with latent confounders. \cite{forney2017counterfactual} circumvented some problems caused by unobserved confounders in multi-armed bandits by using counterfactual-based decision making. \cite{zhang2017transfer} leveraged causal inference to tackle the problem of transferring knowledge across bandit agents. Unlike our method, all these works are restricted to bandits, which corresponds to a simplified RL setting without state transitions.

Last but not least, we want to mention some connections between our method and partially observable Markov decision processes (POMDPs). To the best of our knowledge, existing work on POMDPs has not considered problems with confounders yet. Apart from the confounding part, in POMDPs the observation only provides partial information about the actual state, so the agent in POMDPs does not necessarily know which actual state it is in. By contrast, for convenience in practice, we simplify this setting and assume that the observation provides all the information about the actual state the agent is in but with some noise. Hence we only need to denoise the observation to obtain the actual state. In this sense, a more related work to our model is probably the world model \citep{ha2018world}, because both models used variational inference to estimate the actual state from the observation. If we treat our model as POMDPs with the confounding bias, it will make our model more complicated but it is worth exploring in the future. 

As far as we are concerned, this is the first attempt to build a bridge between confounding and the full RL problem with observational data. 

\section{Conclusion and Future Work}

We have introduced deconfounding reinforcement learning (DRL), a general method for 
addressing reinforcement learning problems with observational data in which hidden confounders affect
observed rewards and actions.
We have used DRL to obtain deconfounding variants of actor-critic methods and showed how these new variants perform better than
the original vanilla algorithms on new confounding benchmark problems.
In the future, we will collaborate with hospitals and apply our approach to real-world medical datasets. We also hope that our work will stimulate further investigation of connections between causal inference and RL.

\bibliography{reference}

\newpage

\begin{appendices}
\section{Variational Lower Bound for $M_{\text{decon}}$} \label{appendix:vlb}
\begin{align*}
&\log p_{\theta}(\vec{x}, \vec{a}, \vec{r}) \\
=& \log \int_{u}\int_{\vec{z}} p_{\theta}(\vec{x}, \vec{a}, \vec{r}, \vec{z}, u) \text{d}\vec{z}\text{d}u \\
\geq & \int_{u}\int_{\vec{z}} q_{\phi}(\vec{z}, u|\vec{x}, \vec{a}, \vec{r})\log \frac{p_{\theta}(\vec{x}, \vec{a}, \vec{r}, \vec{z}, u)}{q_{\phi}(\vec{z}, u|\vec{x}, \vec{a}, \vec{r})} \text{d}\vec{z}\text{d}u \\
= & \int_{u}\int_{\vec{z}} q_{\psi}(u|\vec{x}, \vec{a}, \vec{r})q_{\phi}(\vec{z}|\vec{x}, \vec{a}, \vec{r})\log \frac{p_{\theta}(\vec{x}, \vec{a}, \vec{r}, \vec{z}, u)}{q_{\psi}(u|\vec{x}, \vec{a}, \vec{r})q_{\phi}(\vec{z}|\vec{x}, \vec{a}, \vec{r})} \text{d}\vec{z}\text{d}u \qquad \text{(factorization assumption)}\\
=& \int_{u}\int_{\vec{z}} q_{\psi}(u|\vec{x}, \vec{a}, \vec{r})q_{\phi}(\vec{z}|\vec{x}, \vec{a}, \vec{r}) \\
&\log \frac{p(u)p(z_1)\left[\prod_{t=1}^{T} p(x_t|z_t,u)p(a_t|z_t, u)p(r_{t+1}|z_t, a_t, u)\right]\left[\prod_{t=2}^{T} p(z_t|z_{t-1}, a_{t-1})\right]}{q_{\psi}(u|\vec{x}, \vec{a}, \vec{r})q_{\phi}(\vec{z}|\vec{x}, \vec{a}, \vec{r})} \text{d}\vec{z}\text{d}u \\
=&\int_{u}\int_{\vec{z}}q(u|\vec{x}, \vec{a}, \vec{r})q(z_1|\vec{x}, \vec{a}, \vec{r})\cdots q(z_T|z_{T-1}, \vec{x}, \vec{a}, \vec{r}) \\
&\log \frac{p(u)p(z_1)\prod_{t=1}^{T} p(x_t|z_t,u)p(a_t|z_t, u)p(r_{t+1}|z_t, a_t, u)\prod_{t=2}^{T} p(z_t|z_{t-1}, a_{t-1})}{q(u|\vec{x}, \vec{a}, \vec{r})q(z_1|\vec{x}, \vec{a}, \vec{r})\cdots q(z_T|z_{T-1}, \vec{x}, \vec{a}, \vec{r})} \text{d}\vec{z}\text{d}u \\
=&\sum_{t=1}^T\int_{u}\int_{z_1}\cdots\int_{z_T}q(u|\vec{x}, \vec{a}, \vec{r})q(z_1|\vec{x}, \vec{a}, \vec{r})\cdots q(z_T|z_{T-1}, \vec{x}, \vec{a}, \vec{r}) \\
&\log\left(p(x_t|z_t,u)p(a_t|z_t, u)p(r_{t+1}|z_t, a_t, u)\right) \text{d}\vec{z}\text{d}u \\
&+\int_{u}\int_{z_1}\cdots\int_{z_T}q(u|\vec{x}, \vec{a}, \vec{r})q(z_1|\vec{x}, \vec{a}, \vec{r})\cdots q(z_T|z_{T-1}, \vec{x}, \vec{a}, \vec{r}) \log \frac{p(u)}{q(u|\vec{x}, \vec{a}, \vec{r})} \text{d}\vec{z}\text{d}u \\
&+\int_{u}\int_{z_1}\cdots\int_{z_T}q(u|\vec{x}, \vec{a}, \vec{r})q(z_1|\vec{x}, \vec{a}, \vec{r})\cdots q(z_T|z_{T-1}, \vec{x}, \vec{a}, \vec{r}) \log \frac{p(z_1)}{q(z_1|\vec{x}, \vec{a}, \vec{r})} \text{d}\vec{z}\text{d}u \\
&+ \sum_{t=2}^T\int_{u}\int_{z_1}\cdots\int_{z_T}q(u|\vec{x}, \vec{a}, \vec{r})q(z_1|\vec{x}, \vec{a}, \vec{r})\cdots q(z_T|z_{T-1}, \vec{x}, \vec{a}, \vec{r})\log \frac{p(z_t|z_{t-1}, a_{t-1})}{q(z_t|z_{t-1}, \vec{x}, \vec{a}, \vec{r})} \text{d}\vec{z}\text{d}u \\
=& \sum_{t=1}^T\int_{u}\int_{z_t} q(u|\vec{x}, \vec{a}, \vec{r})q(z_t | z_{t-1}, \vec{x}, \vec{a}, \vec{r})\log\left(p(x_t|z_t,u)p(a_t|z_t, u)p(r_{t+1}|z_t, a_t, u)\right) \text{d}z_t\text{d}u \\
&+ \int_{u} q(u | \vec{x}, \vec{a}, \vec{r})\log\frac{p(u)}{q(u|\vec{x}, \vec{a}, \vec{r})} \text{d}u \\
&+ \int_{z_1} q(z_1 | \vec{x}, \vec{a}, \vec{r})\log\frac{p(z_1)}{q(z_1|\vec{x}, \vec{a}, \vec{r})} \text{d}z_1 \\
&+ \sum_{t=2}^T\int_{z_{t-1}}\int_{z_t}q(z_t|z_{t-1}, \vec{x}, \vec{a}, \vec{r})\log \frac{p(z_t|z_{t-1}, a_{t-1})}{q(z_t|z_{t-1}, \vec{x}, \vec{a}, \vec{r})} \text{d}{z_{t}}\text{d}{z_{t-1}} 
\end{align*}
\begin{align*}
=& \sum_{t=1}^T \underset{\substack{z_t \sim q(z_t|z_{t-1}, \vec{x},\vec{a},\vec{r}) \\ u \sim q(u|\vec{x},\vec{a},\vec{r})}}{\mathbb{E}}
\left[\log p(x_t|z_t,u)+\log p(a_t|z_t, u)+\log p(r_{t+1}|z_t, a_t, u)\right] \nonumber\\
&-\text{KL}\left(q(u|\vec{x},\vec{a},\vec{r}) || p(u)\right) \nonumber \\
&-\text{KL}\left(q(z_1|\vec{x},\vec{a},\vec{r}) || p(z_1)\right)\nonumber\\
&-\sum_{t=2}^T \underset{z_{t-1} \sim q(z_{t-1}|z_{t-2},\vec{x},\vec{a},\vec{r})}{\mathbb{E}} \left[\text{KL}\left(q(z_t|z_{t-1}, \vec{x},\vec{a},\vec{r}) || p(z_t|z_{t-1},a_{t-1})\right)\right].
\end{align*}

\section{Variational Lower Bound for $M_{\text{alt}}$} \label{appendix:vlb_orin}

\begin{align*}
&\log p_{\theta}(\vec{x}, \vec{a}, \vec{r}) \\
=& \log \int_{\vec{z}} p_{\theta}(\vec{x}, \vec{a}, \vec{r}, \vec{z}) \text{d}\vec{z} \\
\geq & \int_{\vec{z}} q_{\phi}(\vec{z}|\vec{x}, \vec{a}, \vec{r})\log \frac{p_{\theta}(\vec{x}, \vec{a}, \vec{r}, \vec{z})}{q_{\phi}(\vec{z}|\vec{x}, \vec{a}, \vec{r})} \text{d}\vec{z} \\
=& \int_{\vec{z}} q_{\phi}(\vec{z}|\vec{x}, \vec{a}, \vec{r}) \log \frac{p(z_1)\left[\prod_{t=1}^{T} p(x_t|z_t)p(a_t|z_t)p(r_{t+1}|z_t, a_t)\right]\left[\prod_{t=2}^{T} p(z_t|z_{t-1}, a_{t-1})\right]}{q_{\phi}(\vec{z}|\vec{x}, \vec{a}, \vec{r})} \text{d}\vec{z} \\
=&\int_{\vec{z}}q(z_1|\vec{x}, \vec{a}, \vec{r})\cdots q(z_T|z_{T-1}, \vec{x}, \vec{a}, \vec{r}) \\
&\log \frac{p(z_1)\prod_{t=1}^{T} p(x_t|z_t)p(a_t|z_t)p(r_{t+1}|z_t, a_t)\prod_{t=2}^{T} p(z_t|z_{t-1}, a_{t-1})}{q(z_1|\vec{x}, \vec{a}, \vec{r})\cdots q(z_T|z_{T-1}, \vec{x}, \vec{a}, \vec{r})} \text{d}\vec{z} \\
=&\sum_{t=1}^T\int_{z_1}\cdots\int_{z_T}q(z_1|\vec{x}, \vec{a}, \vec{r})\cdots q(z_T|z_{T-1}, \vec{x}, \vec{a}, \vec{r}) \log\left(p(x_t|z_t)p(a_t|z_t)p(r_{t+1}|z_t, a_t)\right) \text{d}\vec{z} \\
&+\int_{z_1}\cdots\int_{z_T}q(z_1|\vec{x}, \vec{a}, \vec{r})\cdots q(z_T|z_{T-1}, \vec{x}, \vec{a}, \vec{r}) \log \frac{p(z_1)}{q(z_1|\vec{x}, \vec{a}, \vec{r})} \text{d}\vec{z} \\
&+ \sum_{t=2}^T\int_{z_1}\cdots\int_{z_T}q(z_1|\vec{x}, \vec{a}, \vec{r})\cdots q(z_T|z_{T-1}, \vec{x}, \vec{a}, \vec{r})\log \frac{p(z_t|z_{t-1}, a_{t-1})}{q(z_t|z_{t-1}, \vec{x}, \vec{a}, \vec{r})} \text{d}\vec{z} \\
=& \sum_{t=1}^T\int_{z_t} q(z_t | z_{t-1}, \vec{x}, \vec{a}, \vec{r})\log\left(p(x_t|z_t)p(a_t|z_t)p(r_{t+1}|z_t, a_t)\right) \text{d}z_t \\
&+ \int_{z_1} q(z_1 | \vec{x}, \vec{a}, \vec{r})\log\frac{p(z_1)}{q(z_1|\vec{x}, \vec{a}, \vec{r})} \text{d}z_1 \\
&+ \sum_{t=2}^T\int_{z_{t-1}}\int_{z_t}q(z_t|z_{t-1}, \vec{x}, \vec{a}, \vec{r})\log \frac{p(z_t|z_{t-1}, a_{t-1})}{q(z_t|z_{t-1}, \vec{x}, \vec{a}, \vec{r})} \text{d}{z_{t}}\text{d}{z_{t-1}} \\
=& \sum_{t=1}^T \underset{\substack{z_t \sim q(z_t|z_{t-1}, \vec{x},\vec{a},\vec{r})}}{\mathbb{E}}
\left[\log p(x_t|z_t)+\log p(a_t|z_t)+\log p(r_{t+1}|z_t, a_t)\right] \nonumber\\
&-\text{KL}\left(q(z_1|\vec{x},\vec{a},\vec{r}) || p(z_1)\right)\nonumber\\
&-\sum_{t=2}^T \underset{z_{t-1} \sim q(z_{t-1}|z_{t-2},\vec{x},\vec{a},\vec{r})}{\mathbb{E}} \left[\text{KL}\left(q(z_t|z_{t-1}, \vec{x},\vec{a},\vec{r}) || p(z_t|z_{t-1},a_{t-1})\right)\right].
\end{align*}

\section{Bi-directional LSTM} \label{appendix:lstm}
In our inference model, we use a similar architecture of bi-directional LSTM to that in \citep{krishnan2017structured}. Apart from different inputs, the main difference is to introduce $a_{t-1}$ to computing the combined hidden feature as shown in the following formula.
\begin{figure}[h]
\centering
\includegraphics[width=0.8\linewidth]{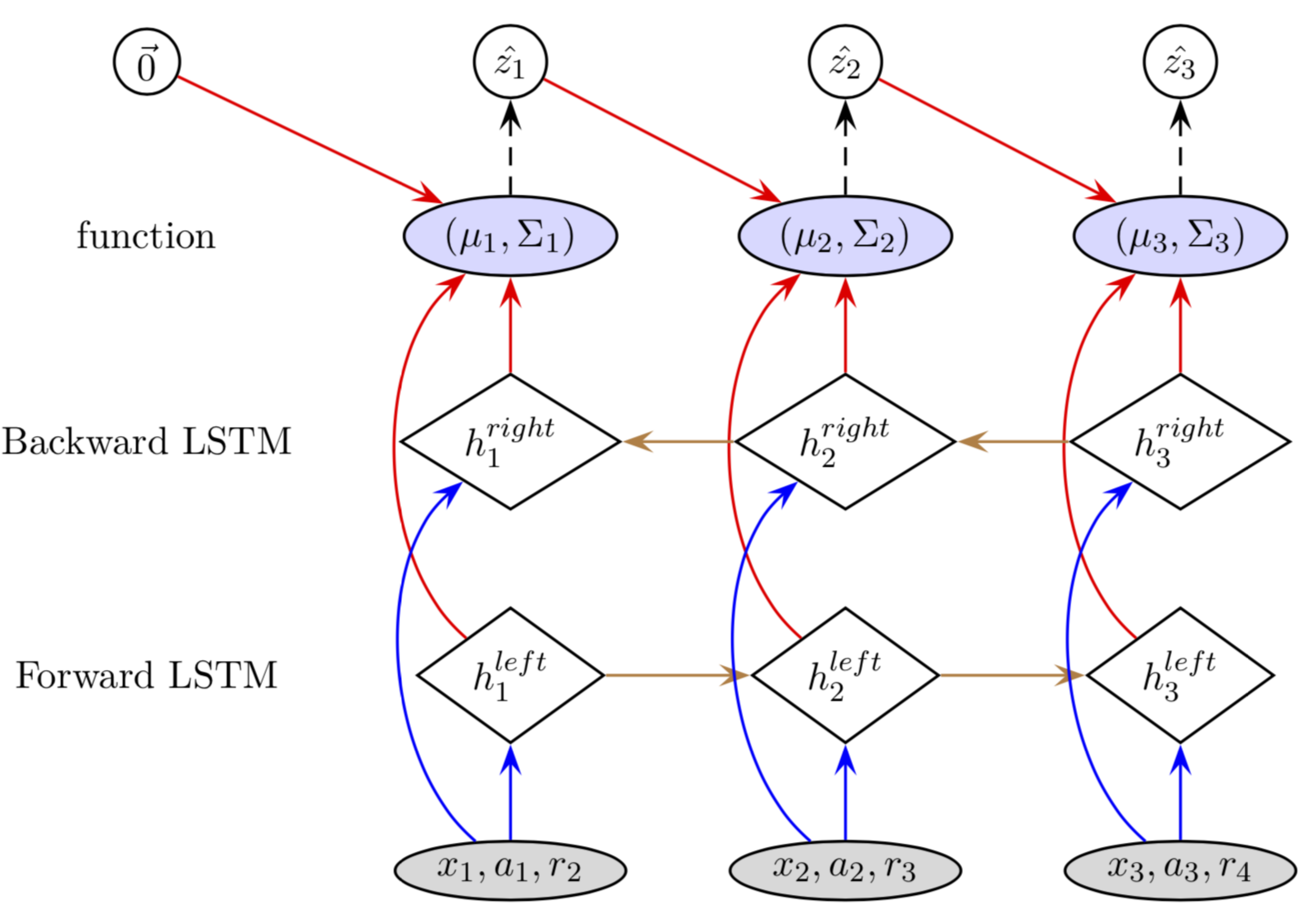}
\end{figure}
\begin{align*}
h_{\text{combined}} &= \frac{1}{4}\left(tanh(W_z \cdot z_{t-1}+b_z) + tanh(W_a \cdot a_{t-1}+b_a) + h_t^{\text{left}} + h_t^{\text{right}}\right) \\
\mu_t &= W_{\mu}\cdot h_{\text{combined}} + b_{\mu} \\
\sigma_t^2 &= \text{softplus}\left(W_{\sigma^2}\cdot h_{\text{combined}} + b_{\sigma^2} \right)
\end{align*}

\section{Example of a proxy variable} \label{appendix:aproxyvariable}

\begin{figure}[h]
\centering
\includegraphics[width=0.25\linewidth]{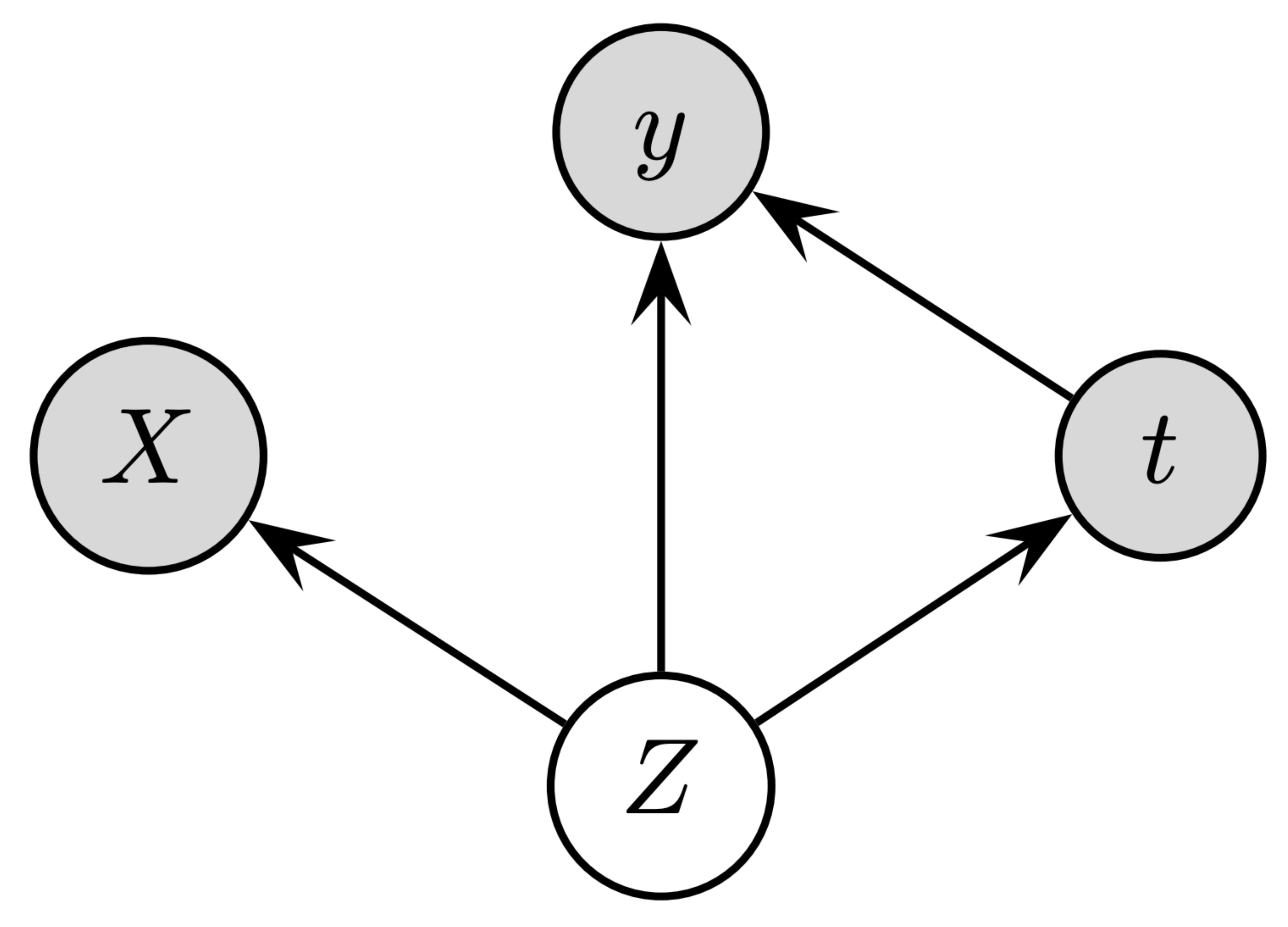}
\caption{Example of a proxy variable. $t$ is a treatment, e.g., medication; $y$ is an outcome, e.g., mortality; $Z$ is an unobserved confounder, e.g., socio-economic status; and $X$ is noisy views on the hidden confounder $Z$, say income in the last year and place of residence.} \label{appendix:fig_proxy}\vspace{-3mm}
\end{figure}

For convenience, we directly move Figure 1 of \citep{louizos2017causal} here, as shown in Figure \ref{appendix:fig_proxy}.

\section{Analysis of the Deconfounding Model} \label{appendix:analysis_deconm}

\begin{figure}[h]
\centering
\includegraphics[width=0.25\linewidth]{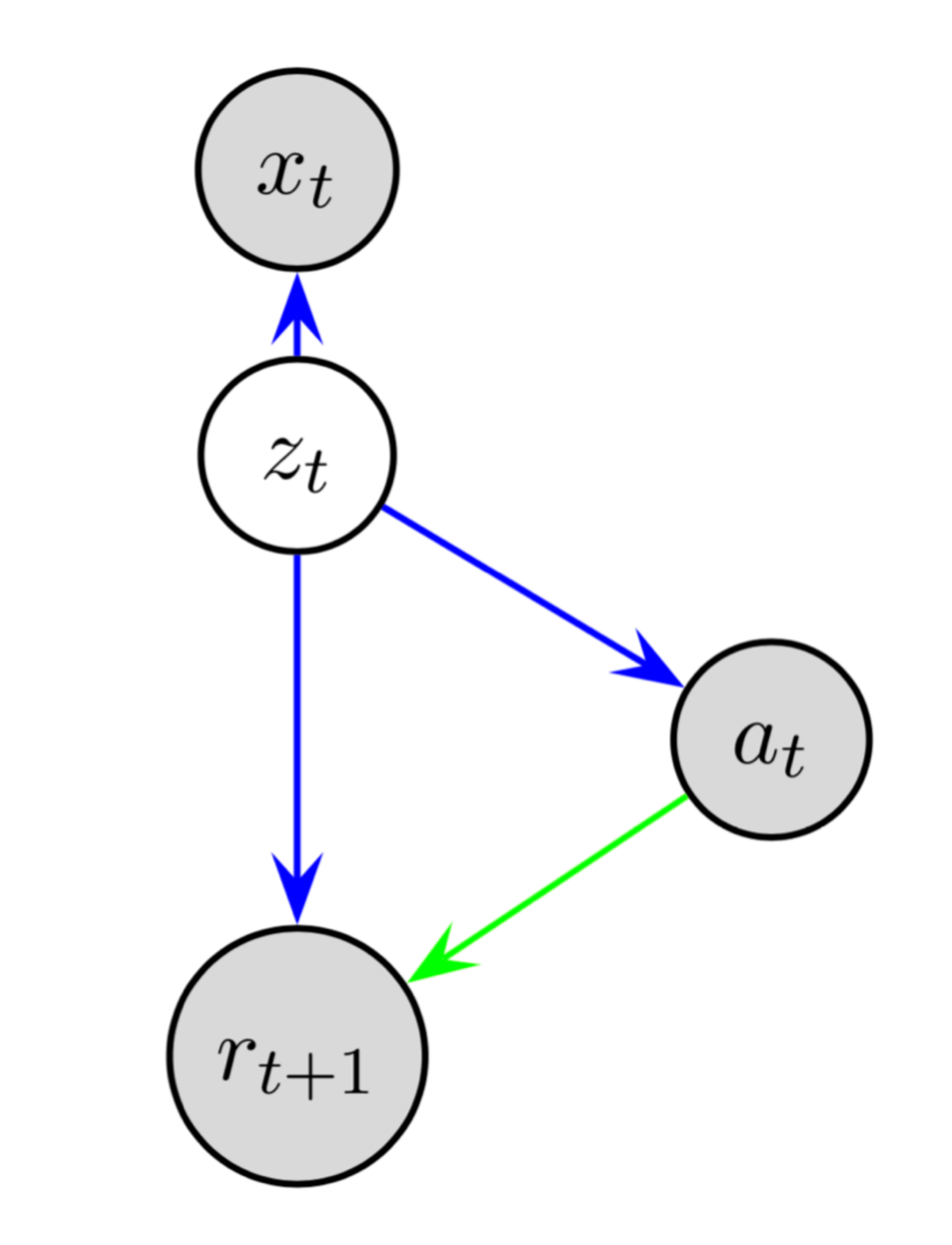}
\caption{The component of 4-tuple $(z_t, x_t, a_t, r_{t+1})$.} \label{appendix:fig_ztuple}
\end{figure}

\begin{figure}[h]
\centering
\includegraphics[width=0.3\linewidth]{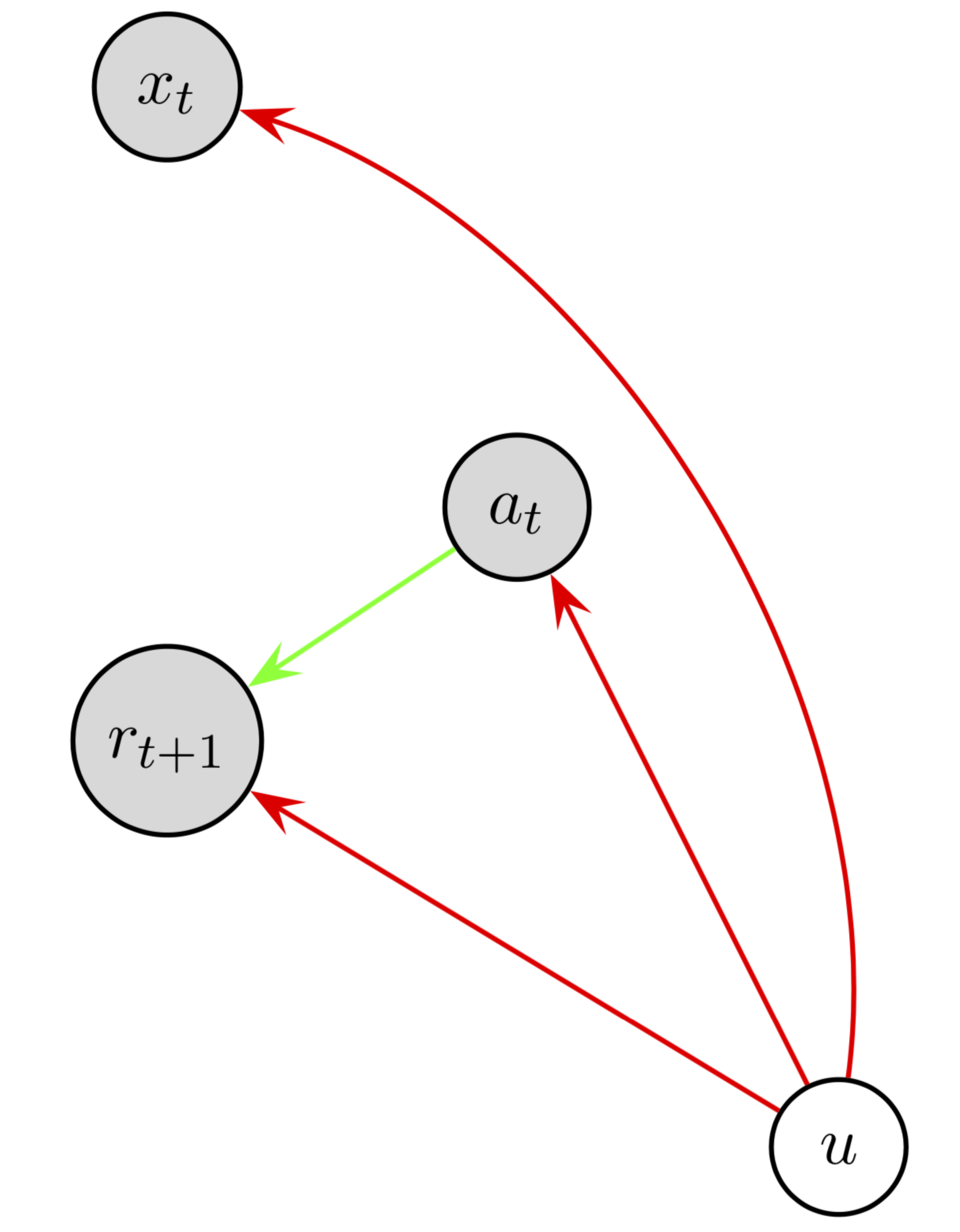}
\caption{The component of 4-tuple $(u, x_t, a_t, r_{t+1})$.} \label{appendix:fig_utuple}
\end{figure}

At each time step $t$, from the deconfounding model presented in Figure \ref{fig:deconRL}, we can extract two 4-tuple components, centering at $z_t$ (Figure \ref{appendix:fig_ztuple}) and $u$ (Figure \ref{appendix:fig_utuple}), respectively. It is apparent to observe that both Figure \ref{appendix:fig_ztuple} and Figure \ref{appendix:fig_utuple} share the exact same structure with Figure \ref{appendix:fig_proxy}.

\section{Confounding Datasets} \label{appendix:confoundingdatasets}

The probabilities between the confounder, action and reward are presented as follows, 
\begin{align}
    &p(u=0) = 0.8, \quad p(u=1)=0.2 ; \label{app_eq_se} \\
    &p(a=T_1 | u=0) = 0.24, \quad p(a=T_2 |u =0) = 0.76; \label{app_eq_poor}  \\
    &p(a=T_1 | u=1) = 0.77, \quad p(a=T_2 |u =1) = 0.23; \label{app_eq_rich} \\
    &p({r_c}=R_1 | a=T_1, u=0) = 0.93, \quad p({r_c}=R_2 | a=T_1, u=0) = 0.07;  \label{app_eq_poor_t1} \\
    &p({r_c}=R_1 | a=T_2, u=0) = 0.87, \quad p({r_c}=R_2 | a=T_2, u=0) = 0.13;  \label{app_eq_poor_t2} \\
    &p({r_c}=R_1 | a=T_1, u=1) = 0.73, \quad p({r_c}=R_2 | a=T_1, u=1) = 0.27;  \label{app_eq_rich_t1} \\
    &p({r_c}=R_1 | a=T_2, u=1) = 0.69, \quad p({r_c}=R_2 | a=T_2, u=1) = 0.31;  \label{app_eq_poor_t2} \\
    & \mu_{{r_c}=R_1} = -1, \quad \mu_{{r_c}=R_2} = -200, \quad \sigma=2. 
\end{align}
Note that, in our model ${r_c}$ is a function of $u$, $a_t$, and $z_t$. We omit the notation of the state in the conditional probability of ${r_c}$ because $z_t$ is fixed at each time step, which does not affect the probability. To help readers understand this setting, we can make a straightforward analogy. More specifically, the confounder $u$ stands for the socio-economic status where $u=0$ represents the poor and $u=1$ the rich. Treatment $T_1$ is more expensive but much better than Treatment $T_2$. $R_1$ means a healthier feedback/recovery than $R_2$. Equation (\ref{app_eq_se}) says that the poor are much more than the rich (it is indeed in reality). Equation (\ref{app_eq_poor}) and Equation (\ref{app_eq_rich}) tell us the fact that the poor tend to choose the cheaper but worse treatment $T_2$ whilst the rich prefer the more expensive but better treatment $T_1$, which also makes sense in the real world. Finally, Equations (\ref{app_eq_poor_t1})-(\ref{app_eq_poor_t2}) reveal that $T_1$ is a better treatment than $T_2$ within both poor and rich population. 

\section{Experimental Settings} \label{appendix:setting}

\begin{table}[h]
\centering
\begin{tabular}{c|c}
\hline
 Hyperparameter  & Value  \\ 
\hline\hline
\multicolumn{2}{c}{\textbf{Deconfounding Model}} \\
\hline
 learning rate & 0.0001 \\ 
 dimension of $z_t$& 50 \\ 
 dimension of $x_t$& 784  \\ 
 dimension of $a_t$& 1  \\ 
 dimension of $r_t$& 1  \\ 
 dimension of $u$& 2  \\ 
 dimension of LSTM unit & 100 \\
 batch size & 128 \\
 number of steps & 5 \\
 number of epoch & 400 \\
 \hline
 \hline
 \multicolumn{2}{c}{\textbf{Deconfounding AC}} \\
 \hline
 number of episodes & 1500 \\
 number of time steps in each episode & 200 \\
 capacity of the replay memory & 100,000 \\
 batch size & 128 \\
 sample size of $u$ & 200 \\
 \hline
\end{tabular}
\caption{Hyperparameters for deconfounding reinforcement learning. Note that for simplicity, in Section \ref{sect:comparerewards} the prior distribution over $u$ is assumed to be a Bernoulli with parameter $p=0.5$ and the dimension of u is set to 1.}
\end{table}

\begin{sidewaystable}[h]
\centering
\begin{tabular}{c|c}
\hline
 Function & Architecture  \\ 
\hline\hline
\multicolumn{2}{c}{\textbf{Deconfounding Model}} \\
\hline
 $f_{1}, f_{2}$ & $\text{FC}_{512} \rightarrow \text{Conv}_{7, 32} \rightarrow \text{Conv}_{14, 16} \rightarrow \text{Conv}_{28, 1} \rightarrow \text{FC}_{784} \rightarrow\{\text{sigmoid}, \text{softplus}\}$ \\ 
 $f_{3}, f_{4}$ & $\{\text{FC}_{100}, \text{FC}_{100}\} \rightarrow \text{FC}_{200} \rightarrow \text{FC}_{200} \rightarrow \text{FC}_{1} \rightarrow\{\tanh, \text{softplus}\}$  \\  
 $f_{5}, f_{6}$ & $\{\text{FC}_{100}, \text{FC}_{100}, \text{FC}_{100}\} \rightarrow \text{FC}_{100} \rightarrow \text{FC}_{100} \rightarrow \text{FC}_{1} \rightarrow\{\text{sigmoid}, \text{softplus}\}$  \\ 
 $f_{7}, f_{8}$ & $\{\text{FC}_{100}, \text{FC}_{100}\} \rightarrow \text{FC}_{100} \rightarrow \text{FC}_{100} \rightarrow \text{FC}_{50} \rightarrow\{\text{None}, \text{softplus}\}$  \\  
 $f_{9}, f_{10}$ & ${ \{\tiny [\text{Conv}_{5,16}\rightarrow \text{Conv}_{5,32}\rightarrow \text{Conv}_{5,32}\rightarrow \text{FC}_{100}], \text{FC}_{100}, \text{FC}_{100}\}\rightarrow \text{FC}_{100}\rightarrow \text{FC}_{100}\rightarrow \text{LSTM}_{100, 5}\rightarrow \text{FC}_{1}\rightarrow\{\text{None}, \text{softplus}\} }$ \\ 
 $f_{11}, f_{12}$ &  ${ \{\tiny [\text{Conv}_{5,16}\rightarrow \text{Conv}_{5,32}\rightarrow \text{Conv}_{5,32}\rightarrow \text{FC}_{100}], \text{FC}_{100}, \text{FC}_{100}\}\rightarrow \text{FC}_{100}\rightarrow \text{FC}_{100}\rightarrow \text{LSTM}_{100, 5}\rightarrow \text{FC}_{50}\rightarrow\{\text{None}, \text{softplus}\} }$ \\ 
 $f_{13}, f_{14}$ & $\text{Conv}_{5,16}\rightarrow \text{Conv}_{5,32}\rightarrow \text{Conv}_{5,32}\rightarrow \text{FC}_{1} \rightarrow\{\tanh, \text{softplus}\}$  \\ 
 $f_{15}, f_{16}$ &  ${ \{[\text{Conv}_{5,16}\rightarrow \text{Conv}_{5,32}\rightarrow \text{Conv}_{5,32}\rightarrow \text{FC}_{100}], \text{FC}_{100}\}\rightarrow \text{FC}_{100}\rightarrow \text{FC}_{1}\rightarrow\{\text{sigmoid}, \text{softplus}\} }$ \\  
 \hline
 \hline
 \multicolumn{2}{c}{\textbf{Deconfounding AC}} \\
 \hline
 $V(z_t;\phi_V)$ & $\text{FC}_{300} \rightarrow \text{FC}_{300} \rightarrow \{\text{None}, \text{softplus}\}$ \\
 $\pi(a_t | z_t; \theta)$ & $\text{FC}_{300} \rightarrow \text{FC}_{300} \rightarrow \{\tanh, \text{softplus}\}$ \\
 \hline
\end{tabular}
\caption{Architectures for deconfounding reinforcement learning. Here $\text{FC}_{k}$ stands for a fully-connected layer with $k$ units, $\text{Conv}_{k,n}$ for a convolution layer with $n$ filters of size $k \times k$, $\text{LSTM}_{n, t}$ for a LSTM layer rolling out for $t$ steps with latent size of $n$, $\{\cdot\}$ for the parallel operators, and $[\cdot]$ for the sequential operators. In default, $\text{FC}_{k}$ and $\text{Conv}_{k,n}$ are followed by a softplus activation layer and a batch-norm layer, which are omitted here for simplicity. Note that, in our setting, the two functions in each pair $\{f_{2i-1}, f_{2i}\}_{i=1,\ldots,8}$ share the same parameters except for the last layer.}
\end{sidewaystable}

\end{appendices}

\end{document}